\documentclass[10pt,twocolumn,letterpaper]{article}

\usepackage[pagenumbers]{cvpr} 
\usepackage{algorithm}
\usepackage{algpseudocode}
\usepackage{multirow}
\definecolor{cvprblue}{rgb}{0.21,0.49,0.74}
\usepackage[pagebackref,breaklinks,colorlinks,allcolors=cvprblue]{hyperref}

\def\paperID{*****} 
\def\confName{CVPR}
\def\confYear{2026}

\title{MarKey: Marginal Utility Guided Greedy Keyframe Selection for Long Video Understanding}

\author{
Hongchang Shi$^{1}$ \quad
Jinpeng Hu$^{1}$ \quad
Ao Wang$^{1}$ \quad
Wenzheng Zhou$^{1}$ \quad
Hui Ma$^{1}$ \quad
Feng Li$^{1}$ \quad
Zenglin Shi$^{1}$\\
$^{1}$Hefei University of Technology, Hefei, China\\
{\tt\small 2024170833@mail.hfut.edu.cn, 135858hjp@gmail.com}
}

\begin{document}
\maketitle
\begin{abstract}
Long-video understanding remains challenging for multimodal large language models (MLLMs) because densely encoding long frame sequences is computationally expensive, while uniform sampling under a limited visual budget can miss sparse yet decisive evidence.
Recent training-free keyframe selection methods have enabled more efficient inference and yielded promising performance gains.
However, many existing methods score frames largely in isolation without explicitly considering how each candidate complements the currently selected subset, potentially resulting in redundant selections and incomplete evidence coverage.
To address this limitation, we propose MarKey, a training-free framework that formulates keyframe selection as subset-aware greedy optimization. 
At each iteration, MarKey scores each candidate using a tractable surrogate that jointly accounts for query relevance, marginal coverage gain, and context-dependent redundancy, and selects the frame with the highest utility.
To make this iterative subset-aware evaluation efficient, MarKey uses a compact set of representative anchors to approximate full-video coverage and a bounded window of previously selected frames to limit context-dependent comparisons.
Experiments on six benchmarks spanning holistic video understanding, human-centric video understanding, and open-ended video understanding demonstrate that MarKey consistently outperforms existing methods.
Further analyses show robust gains across different MLLM backbones, model scales, and frame budgets.
\end{abstract}
\section{Introduction}

Multimodal large language models (MLLMs) have demonstrated strong capabilities in visual understanding and reasoning across a wide range of image and video tasks 
\citep{alayrac2022flamingo,li2023blip,ye2024mplug}.
As these models are increasingly applied to more complex video scenarios, long-form video understanding has emerged as an important frontier. 
This progress is accompanied by a growing body of benchmarks that demand multi-step reasoning \citep{lei2018tvqa,xiao2021next}, long-horizon evidence aggregation \citep{mangalam2023egoschema,wang2025lvbench}, comprehensive reasoning over complex real-world scenarios \citep{hu2025beyond,hu2026mmhbench,hu2025emobench}, and fine-grained temporal understanding \citep{liu2024tempcompass,cai2024temporalbench} over videos spanning minutes or even hours.
\begin{figure}[t]
\centering
\includegraphics[width=0.480\textwidth, trim=0 0 0 0]{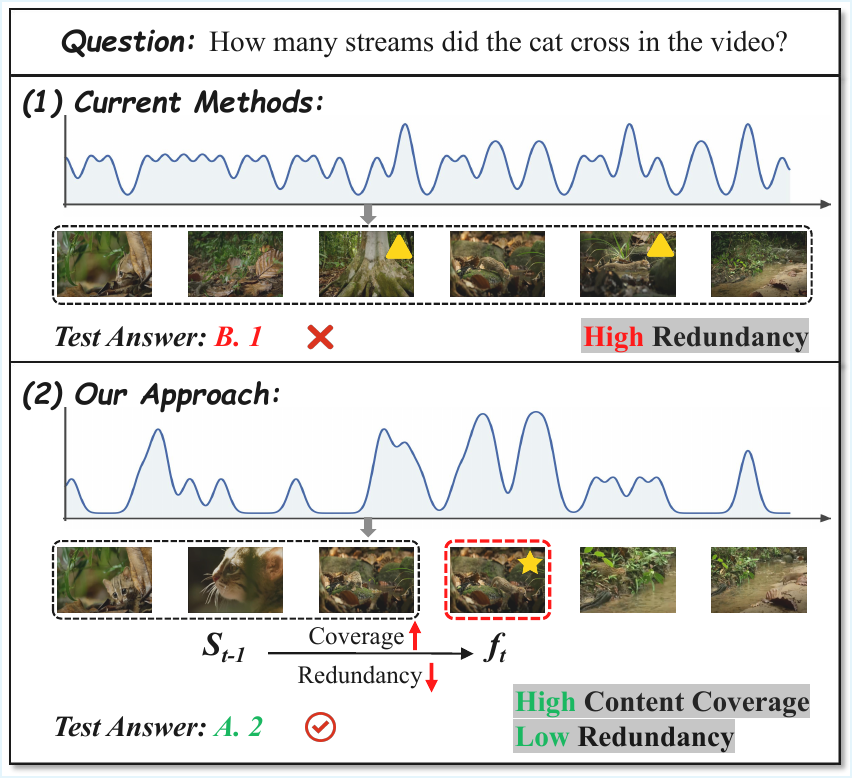}
\caption{Existing methods often select redundant frames around salient moments, limiting content coverage. Our query-guided selection balances relevance, coverage, and redundancy to yield compact, complementary keyframes.}
\label{fig:question_responder_role}
\vspace{-2em}
\end{figure}
Compared with short videos, long-video understanding poses a more fundamental evidence-allocation challenge, as task-relevant cues are often sparse, temporally dispersed, and meaningful only when interpreted in relation to the query and the broader video narrative.
Because densely encoding an entire long video is computationally prohibitive, most MLLMs operate under a fixed visual budget and process only a limited number of uniformly sampled frames.
Although simple and efficient, uniform sampling provides increasingly coarse temporal coverage as video duration grows, making it more prone to miss brief yet decisive events required for downstream reasoning.
These limitations have motivated frame selection methods that allocate the constrained visual budget more effectively before costly MLLM inference.
Existing approaches can be broadly categorized into learning-based and training-free methods. 
Learning-based methods explicitly optimize a querying or ranking policy to select frames that are useful for Video-LLMs, often relying on additional supervision or proxy objectives \citep{yu2023self,hu2025m}. 
For example, ReFoCUS \citep{lee2025refocus} treats selection as an autoregressive decision policy and applies reinforcement learning with rewards derived from a frozen reference multimodal model to directly optimize evidence selection for temporally grounded reasoning.
Although effective, such methods typically require extra training and may be less convenient to transfer across different MLLM backbones, prompts, and deployment settings. 
Training-free methods instead construct frame subsets at inference time using pretrained vision-language representations \citep{sun2025mdp3,liu2025bolt,tang2025adaptive,zhu2025focus}.
Their plug-and-play nature makes them particularly attractive for long-video understanding across heterogeneous MLLM backbones.
For instance, AKS \citep{tang2025adaptive} introduces a keyframe selection method that recursively partitions the video and adaptively selects keyframes.
FOCUS \citep{zhu2025focus} proposes a budgeted evidence search strategy that progressively identifies informative temporal regions and selects query-relevant frames under a strict frame budget.
Despite this progress, a fundamental question remains insufficiently addressed: \emph{how much new evidence does a candidate frame contribute beyond the frames already selected?}
Under a strict frame budget, individual relevance and subset utility are not equivalent.
As illustrated in Fig.~\ref{fig:question_responder_role}, multiple frames depicting the same stream-crossing event may all appear highly relevant to the question.
However, once one such frame has been selected, additional visually similar frames contribute little new evidence toward determining the total number of crossings, whereas a moderately less relevant frame may become crucial if it captures a different, previously unobserved event.
The utility of a candidate is therefore conditioned on the evolving selection context, since a frame that is informative in isolation may become redundant after similar evidence has been included.
This context dependence exposes a mismatch between frame-level scoring and the objective of keyframe selection.
Although recent methods employ sequential selection, recursive partitioning, or heuristic search, many do not explicitly optimize this subset-conditioned marginal contribution.
Consequently, the selected frames may be individually relevant yet collectively redundant, resulting in incomplete evidence coverage.
This problem becomes increasingly severe as video duration grows because more potentially relevant events must compete for the same fixed number of visual slots.

To address this limitation, keyframe selection should move beyond local relevance scoring and evaluate each candidate relative to the evolving subset.
Accordingly, we propose \textbf{MarKey}, a training-free framework that reformulates keyframe selection from independent frame prioritization into greedy optimization of subset-conditioned marginal utility.
At each iteration, MarKey estimates the additional evidence contributed by each remaining candidate by jointly considering query relevance, anchor-based marginal coverage gain, and context-dependent redundancy.
Query relevance captures the candidate's alignment with the question, marginal coverage gain measures its contribution to previously underrepresented video content, and the redundancy penalty suppresses substantial overlap with the current selection.
Together, these complementary signals provide a tractable surrogate for the candidate's marginal contribution, enabling MarKey to favor frames that are both query-relevant and evidentially complementary.
MarKey then adopts a greedy selection strategy that iteratively selects the frame with the highest surrogate utility, yielding a compact yet informative subset for downstream reasoning.
Representative anchors approximate video-wide coverage, while a bounded active context limits redundancy comparisons over the selection history, making iterative subset-aware evaluation computationally practical.
We evaluate MarKey on six benchmarks spanning holistic, human-centric, and open-ended video understanding.
Further analyses demonstrate robust gains across MLLM backbones, model scales, and frame budgets. 
Component ablations and selected-frame redundancy analysis further verify the complementary effects of relevance, coverage, and redundancy modeling.

\vspace{-2pt}
Our main contributions are summarized as follows:
\begin{itemize}
\item We formulate training-free keyframe selection as a dynamic subset-construction problem, shifting candidate evaluation from standalone relevance to the additional evidence contributed beyond the selected subset.

\item We propose MarKey, whose surrogate utility jointly models query relevance, video-wide coverage, and context-dependent redundancy to construct complementary evidence subsets.

\item We make iterative subset-aware selection efficient by using representative anchors to approximate video-wide coverage and a bounded active context to limit redundancy comparisons.

\item Extensive experiments on six benchmarks demonstrate consistent improvements. Further analyses confirm robustness across MLLM backbones, model scales, frame budgets, and video durations.

\end{itemize}

\section{Related Work}
\subsection{MLLMs for Long-Video Understanding}
MLLMs build upon recent advances in LLMs, which have substantially improved natural language understanding and reasoning across a broad range of tasks \citep{dai2026psyche,xu2025multiagentesc,hu2026agentmental}. By incorporating visual encoders and unified token interfaces, MLLMs further extend these capabilities to joint reasoning over language and visual inputs.
Early research in this area primarily focused on image-text understanding tasks \citep{zhang2023llavar,zhao2023svit}. Representative methods such as LLaVA \citep{liu2023visual} and MiniGPT-4 \citep{zhu2023minigpt} adopt a modular design, where a pretrained visual encoder extracts image features and a lightweight projection module aligns them with the language space of a large language model.
Furthermore, many recent studies have begun to extend multimodal large language models from static images to video inputs by encoding sampled frames and integrating temporal visual information. VideoChat \citep{li2025videochat}, Video-ChatGPT \citep{maaz2024video}, Video-LLaVA \citep{lin2024video} and Video-LLaMA \citep{zhang2023video} follow this paradigm and employ multimodal instruction tuning to enable large language models to reason over short video clips.
As research progresses, a series of architectural refinements has been proposed to better model temporal information and improve the efficiency of visual token processing. Models like LLaVA-OneVision \citep{li2024llava}, LLaVA-NeXT \citep{li2024llava1} and its video variants, Aria \citep{li2024aria}, PLLaVA \citep{xu2024pllava} and Kangaroo \citep{liu2024kangaroo} unify multi-granularity visual inputs, strengthen temporal adapters and refine projection modules or training curricula.
Recent work further shifts attention to long videos and extended visual contexts. Works such as LongVILA \citep{chen2024longvila}, LongVA \citep{zhang2024long} and LongVLM \citep{weng2024longvlm} extend the effective context length of large language models and introduce hierarchical or multi-level representations, enabling more robust reasoning over long untrimmed video sequences.

\subsection{Efficient Long-Video Understanding}
The computational burden of long-video MLLMs is largely caused by the rapid accumulation of visual tokens, which increases both multimodal encoding cost and the context consumed during language-model inference.
Beyond directly extending the context window, a growing body of research improves efficiency by redesigning how long visual streams are represented, compressed, and accessed.
One line of work constructs compact visual representations within the model \citep{lan2024vidcompress,huang2025prunevid,li2024vidtome}.
LLaMA-VID \citep{li2024llama} represents each frame using a small number of content and context tokens, substantially reducing the visual sequence length while retaining frame-specific information.
Video-XL \citep{shu2025video} exploits key-value sparsification and dynamic compression to summarize visual information over long temporal intervals, whereas LongVU \citep{shen2024longvu} adaptively removes spatial and temporal redundancy according to inter-frame dependencies and textual guidance.
Another line of research avoids encoding the entire video into a single dense context and instead organizes visual evidence through structured abstraction and retrieval \citep{wang2025videotree,yuan2025memory,ma2025drvideo}.
For example, Video-RAG \citep{luo2026video} builds an auxiliary multimodal knowledge base from long-video content and retrieves question-relevant visual and textual evidence before MLLM inference, enabling the reasoning model to access a compact context without processing the complete video sequence.
While effective, these approaches often rely on specialized compression modules, external memory, or auxiliary retrieval pipelines, which introduce additional computational and system overhead and may discard fine-grained visual evidence during abstraction.
This motivates lightweight pre-processing strategies, particularly keyframe selection, as a practical and complementary solution for reducing visual redundancy while preserving task-relevant evidence in its original form.

\subsection{Keyframe Selection for Long Videos} Keyframe selection has attracted considerable attention in efficient video understanding, as it aims to identify a compact yet informative subset of frames that preserves salient visual content for downstream reasoning.
Early efforts mainly focus on task-supervised selector learning for conventional video models, where the selector is optimized together with downstream recognition objectives \citep{wu2019adaframe,wu2019multi,korbar2019scsampler,gowda2021smart,zhao2023search}.
More recently, with the rise of MLLMs, training-based keyframe selection has evolved into a new paradigm that leverages supervision distilled from large models rather than relying solely on task labels. 
Representative works include Frame-Voyager \citep{yu2024frame}, which trains a query-aware frame selector under the supervision of a pretrained MLLM, and M-LLM video frame selector \citep{hu2025m}, which supervises a lightweight selector with MLLM-derived single-frame relevance and multi-frame complementarity signals.
In parallel, training-free methods have recently gained substantial attention, as they avoid additional selector training and can be readily integrated with frozen MLLMs. These approaches typically construct frame subsets at inference time using pretrained vision-language representations together with lightweight scoring strategies.
Along this line, MDP3 \citep{sun2025mdp3} formulates frame selection as a query-conditioned list-wise sequential selection problem, BOLT \citep{liu2025bolt} explores inference-time query-guided frame selection, and FOCUS \citep{zhu2025focus} proposes a training-free exploration-based strategy that progressively locates informative regions before selecting keyframes.
However, most existing methods still emphasize frame-wise relevance, often failing to preserve the semantic completeness of the selected subset.
Our work, MarKey, addresses this issue by selecting frames based on context-aware marginal utility, yielding subsets that are more complete and less redundant.

\section{Methods}

\subsection{Problem Formulation}
\textbf{Keyframe Selection.} 
\begin{figure*}[t]
\centering
\includegraphics[width=0.99\textwidth, trim=0 0 0 0]{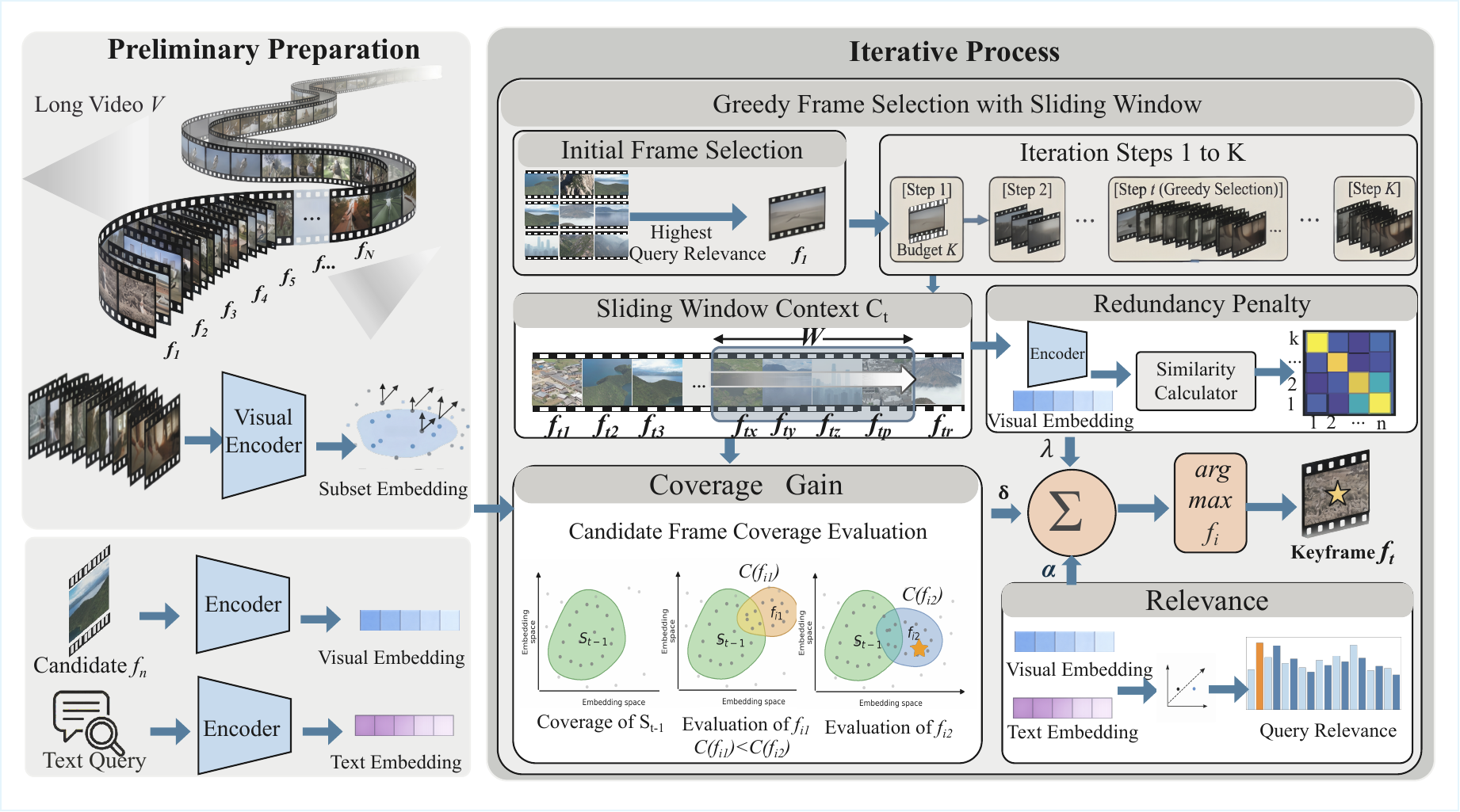}
\vspace{-1em}
\caption{An overview of our proposed MarKey framework. The pipeline consists of two main stages: (1) \textbf{Preliminary Preparation}: Given a long video $\mathcal{V}$ and a text query $q$, we first construct a representative anchor subset to summarize the global visual content of the video, and encode the anchor subset, candidate frames, and the query into a shared embedding space. (2) \textbf{Greedy Frame Selection with Sliding Window}: at each step, a bounded context window is formed from previously selected frames, and every remaining candidate is evaluated by a contextual utility that combines query relevance, coverage gain, and redundancy penalty. These three terms are weighted and aggregated, and the candidate with the maximum utility is selected as the next keyframe. Repeating this process until the frame budget K is reached yields a compact subset that is query-relevant and visually representative.}
\label{fig:main}
\end{figure*}
MLLMs are fundamentally constrained by a limited visual context budget, which makes it impractical to process all frames of a long video simultaneously.
For long-form videos, directly encoding the full visual stream incurs prohibitive token and memory costs, while naive temporal subsampling may miss query-relevant moments that are sparse in time but crucial for downstream reasoning.
Therefore, long-video understanding typically requires selecting a compact subset of frames that preserves the visual evidence most useful for answering the query.
Formally, given a long video $\mathcal{V} = (f_1, f_2, \dots, f_N)$ with $N$ temporally ordered frames and a textual query $q$, the goal of query-guided keyframe selection is to choose a compact subset of $K$ frames, such that $\mathcal{S}$ preserves the visual evidence most relevant to downstream reasoning. The selected subset is expected to form a compact and informative representation of the original video under a strict frame budget.
\begin{equation}
\mathcal{S} = \{f_{i_1}, f_{i_2}, \dots, f_{i_K}\} \subseteq \mathcal{V}, \qquad 1 \leq K \ll N.
\label{eq:keyframe_selection}
\end{equation}
\noindent\textbf{Optimization Objective.}
Given the candidate frame set defined above, our goal is to select a subset of $K$ frames that maximizes its utility for downstream query-guided reasoning. We formulate query-guided keyframe selection as the following subset optimization problem:
\begin{equation}
\mathcal{S}^{*}
=
\arg\max_{\mathcal{S} \subseteq \mathcal{V},\, |\mathcal{S}| = K}
U(\mathcal{S}; q),
\label{eq:subset_objective}
\end{equation}
where $U(\mathcal{S}; q)$ measures the overall quality of a selected frame subset for answering the query. 

\subsection{Context-Aware Marginal Utility}
To optimize the subset objective in Eq.~\eqref{eq:subset_objective}, we construct the selected subset progressively and evaluate each candidate frame by the additional utility it contributes to the current partial selection.
Formally, given the partial selection $\mathcal{S}_{t-1}$ at step $t$, the ideal contextual contribution of candidate frame $f_i$ is defined as
\begin{equation}
\Delta_i^{(t)}
=
U(\mathcal{S}_{t-1}\cup\{f_i\}; q)
-
U(\mathcal{S}_{t-1}; q).
\label{eq:marginal_gain}
\end{equation}
This context-dependent formulation encourages the selection of frames that provide complementary evidence beyond the current subset, instead of repeatedly favoring individually relevant but redundant frames.
Directly optimizing Eq.~\eqref{eq:marginal_gain} is computationally expensive for long videos, because it requires repeatedly evaluating subset-level utility during selection.
We therefore approximate this contextual marginal utility with a tractable surrogate composed of three complementary terms: query relevance, coverage gain, and redundancy suppression.
Together, these terms encourage the selector to retain frames that are relevant to the query, improve the representational completeness of the current subset, and avoid repeated evidence.

\noindent\textbf{Query Relevance.}
To quantify the semantic relevance of each frame to the query, a pretrained vision-language encoder, such as CLIP \citep{radford2021learning}, is used to extract the visual embedding of each candidate frame and the textual embedding of the query.
Let $\mathbf g_i$ denote the visual embedding of frame $f_i$, and let $\mathbf q$ denote the textual embedding of query $q$. After $\ell_2$ normalization, the query relevance score is computed as
\begin{equation}
r_i
=
\cos(\mathbf g_i,\mathbf q)
=
\frac{\mathbf g_i^\top \mathbf q}
{\|\mathbf g_i\|_2\,\|\mathbf q\|_2},
\qquad i=1,\ldots,N.
\label{eq:query_relevance}
\end{equation}
This term serves as the primary query-conditioned matching signal, encouraging the selector to retain frames that are semantically aligned with the user’s intent.

\noindent  \textbf{Coverage Gain.}
A high-quality keyframe subset should not only be relevant to the query, but also preserve sufficiently broad visual evidence from the original video.
In other words, the selected frames should provide strong representational coverage, so that the final subset can serve as a compact yet faithful surrogate of the full video for downstream reasoning.
However, many existing query-guided keyframe selection methods primarily emphasize frame-query relevance, while paying limited attention to whether the selected frames collectively cover the broader visual content of the video.
As a result, the resulting subset may be locally relevant but globally incomplete, leaving potentially useful visual evidence insufficiently represented.

To address this limitation, we explicitly model the marginal coverage gain of each candidate with respect to the current partial selection.
The goal is to measure how much additional visual coverage a candidate provides beyond what has already been captured by the selected subset.
A natural formulation is to define this contribution over the full candidate set.
Specifically, let $s_{uj}$ denote the cosine similarity between frames $f_u$ and $f_j$ computed from normalized global visual embeddings, where $u$ indexes a frame in the full candidate pool and $j$ indexes a selected frame in the current subset $\mathcal{S}_{t-1}$.
The ideal representational coverage of a selected subset $\mathcal{S}_{t-1}$ is defined as
\begin{equation}
F_{\mathrm{full}}(\mathcal{S}_{t-1})
=
\frac{1}{N}\sum_{u=1}^{N}\max_{j\in\mathcal{S}_{t-1}} s_{uj},
\label{eq:full_coverage}
\end{equation}
which measures how well the frames in $\mathcal{S}_{t-1}$ collectively represent the visual content of the entire candidate pool.

From a set-level perspective, the term $\max_{j\in\mathcal{S}_{t-1}} s_{uj}$ indicates how well frame $f_u$ is covered by the selected subset $\mathcal{S}_{t-1}$.
Thus, Eq.~\eqref{eq:full_coverage} evaluates the representational completeness of the selected subset as a whole, rather than the quality of individual frames in isolation.
Given the current partial selection $\mathcal{S}_{t-1}$, we define the current coverage state of frame $f_u$ as
\begin{equation}
m_u^{(t-1)}=
\begin{cases}
0, & \mathcal{S}_{t-1}=\emptyset,\\[4pt]
\max_{j\in\mathcal{S}_{t-1}} s_{uj}, & \text{otherwise},
\end{cases}
\label{eq:coverage_state}
\end{equation}
which measures the degree to which $f_u$ has already been represented by the currently selected subset.
Under this formulation, the ideal marginal coverage contribution of candidate frame $f_i$ is
\begin{equation}
\kappa^{(t)}_{i,\mathrm{full}}
=
F_{\mathrm{full}}(\mathcal{S}_{t-1}\cup\{f_i\})
-
F_{\mathrm{full}}(\mathcal{S}_{t-1}).
\label{eq:full_coverage_gain}
\end{equation}
Equivalently, it can be written as
\begin{equation}
\kappa^{(t)}_{i,\mathrm{full}}
=
\frac{1}{N}\sum_{u=1}^{N}
\max\!\bigl(s_{ui}-m_u^{(t-1)},\,0\bigr),
\label{eq:full_coverage_gain_equiv}
\end{equation}
which explicitly quantifies how much additional coverage candidate frame $f_i$ contributes beyond the current selection context. 
%

\noindent  \textbf{Redundancy Penalty.}
Coverage gain rewards candidates that complement the current subset, but it does not explicitly discourage repeated evidence. This limitation is particularly important in long videos, where temporally adjacent segments, repetitive scenes, and recurring viewpoints often produce highly similar frames.
Under a strict frame budget, a candidate frame is of limited utility when its content largely overlaps with the current subset and offers little additional evidence for query-guided reasoning, even if it is individually highly relevant.
To explicitly suppress repeated evidence, we introduce a context-dependent redundancy penalty.
Unlike coverage gain, which rewards candidates for improving subset completeness, the redundancy term penalizes candidates whose content is already well represented by the current selection context.

Formally, given the partial selection $\mathcal{S}_{t-1}$ at step $t$, we define the redundancy penalty of candidate frame $f_i$ as
\begin{equation}
\rho_i^{(t)} =
\begin{cases}
0, & \mathcal{S}_{t-1} = \emptyset, \\[4pt]
\max_{j \in \mathcal{S}_{t-1}} \cos(\mathbf g_i,\mathbf g_j), & \text{otherwise},
\end{cases}
\label{eq:redu}
\end{equation}
where $\cos(\cdot,\cdot)$ denotes cosine similarity between normalized global visual embeddings. 
A larger value of $\rho_i^{(t)}$ indicates that candidate frame $f_i$ overlaps more strongly with the current subset and is therefore less likely to contribute new evidence.
By penalizing such overlap, this term suppresses redundant selections and encourages a more compact and informative keyframe subset.

\noindent\textbf{Surrogate Marginal Utility.}
Following the contextual marginal-utility formulation in Eq.~\eqref{eq:marginal_gain}, we approximate the marginal contribution of each candidate frame with a tractable surrogate that combines query relevance, anchor-based coverage gain, and redundancy suppression. Specifically, for candidate frame $f_i$ at step $t$, we define the surrogate marginal utility as
\begin{equation}
u^{(t)}(f_i)
=
\alpha r_i
+
\delta \kappa_i^{(t)}
-
\lambda \rho_i^{(t)},
\label{eq:overall_utility}
\end{equation}
where $\alpha$, $\delta$, and $\lambda$ balance the contributions of query relevance, coverage improvement, and redundancy suppression, respectively. Eq.~\eqref{eq:overall_utility} serves as a tractable estimate of the ideal marginal utility of frame $f_i$ under the current partial selection.
\subsection{Optimization and Greedy Selection}
\begin{algorithm}[t]
\caption{Greedy Keyframe Selection with Anchor-Based Coverage}
\label{alg:selection}
\small
\begin{algorithmic}[1]
\Require Candidate frame set $\mathcal{V}=\{f_i\}_{i=1}^{N}$, query $q$, frame budget $K$, representative anchor subset $\mathcal{A}$, window size $W$
\Ensure Selected subset $\mathcal{S}_K$

\State Encode candidate frames $\{f_i\}_{i=1}^{N}$ into normalized visual embeddings $\{\mathbf g_i\}_{i=1}^{N}$, and encode query $q$ into $\mathbf q$
\State Compute query relevance scores $r_i=\cos(\mathbf g_i,\mathbf q)$ for all candidates
\State Compute pairwise similarities $s_{ui}$ between anchor frames $f_u\in\mathcal{A}$ and candidate frames $f_i\in\mathcal{V}$
\State Initialize $\mathcal{S}_0\gets\emptyset$ and $m_u^{(0)}\gets 0$ for all $u\in\mathcal{A}$

\For{$t=1$ to $K$}
    \State Construct active context window $\mathcal{C}_{t-1}\subseteq\mathcal{S}_{t-1}$ as the context for evaluation
    \For{each candidate $f_i \in \mathcal{V}\setminus\mathcal{S}_{t-1}$}
        \State Compute redundancy penalty $\rho_i^{(t)}$ by Eq.~\eqref{eq:redu}
         \State Compute anchor-based coverage gain $\kappa_i^{(t)}$ by Eq.~\eqref{eq:anchor_coverage_gain}
              \State Compute surrogate marginal utility $u_i^{(t)}$ by Eq.~\eqref{eq:overall_utility}
    \EndFor
    \State Select
    \[
    f_{i_t}^{*}
=
\arg\max_{f_i\in\mathcal{V}\setminus\mathcal{S}_{t-1}}
u^{(t)}(f_i).
    \]
    \State Update selected subset
    \[
    \mathcal{S}_t \gets \mathcal{S}_{t-1}\cup\{f_{i_t}^{*}\}
    \]
    \For{each anchor $f_u\in\mathcal{A}$}
        \State Update anchor coverage state
        \[
        m_u^{(t)} \gets \max\!\bigl(m_u^{(t-1)},\, s_{u\,i_t^{*}}\bigr)
        \]
    \EndFor
\EndFor

\State \Return $\mathcal{S}_K$
\end{algorithmic}
\end{algorithm}
\noindent\textbf{Coverage Gain Optimization.}
Directly evaluating Eq.~\eqref{eq:full_coverage_gain_equiv} over the full candidate pool is computationally expensive for long videos, since the marginal coverage gain of every remaining candidate must be repeatedly computed during greedy selection.
To make this computation tractable, we approximate the full candidate set by a much smaller representative anchor subset $\mathcal{A}\subseteq\mathcal{V}$, where $|\mathcal{A}| \ll |\mathcal{V}|$.
Following the temporal locality observation that nearby video
frames often exhibit strong semantic similarity
\citep{li2023tcovis}, we construct $\mathcal{A}$ through
temporal-relevance stratified sampling.
Specifically, each video is divided into non-overlapping
10-second clips, and frames are uniformly sampled at 1 FPS
within each clip. For every clip, the sampled frames are ranked according to their
query-frame similarity scores and partitioned into high-, medium-, and low-relevance strata.
Anchors are then sampled from all three strata under a fixed
global anchor budget, and the clip-level anchors are merged
across the entire video to form the final anchor subset
$\mathcal{A}$.
In our implementation, the anchor budget is set to 648 frames;
when fewer frames are available, all eligible frames are retained.
Under this approximation, the anchor-based coverage objective is defined as
\begin{equation}
F(\mathcal{S}_{t-1})
=
\frac{1}{|\mathcal{A}|}\sum_{u\in\mathcal{A}}\max_{j\in\mathcal{S}_{t-1}} s_{uj},
\label{eq:anchor_coverage}
\end{equation}
and the corresponding anchor-based marginal coverage gain of candidate frame $f_i$ becomes
\begin{equation}
\kappa_i^{(t)}
=
\frac{1}{|\mathcal{A}|}\sum_{u\in\mathcal{A}}
\max\!\bigl(s_{ui}-m_u^{(t-1)},\,0\bigr).
\label{eq:anchor_coverage_gain}
\end{equation}
Here, $m_u^{(t-1)}$ is defined in the same way as Eq.~\eqref{eq:coverage_state}, except that $u$ is restricted to the anchor subset $\mathcal{A}$.
This anchor-based formulation preserves the set-level representational objective of Eq.~\eqref{eq:full_coverage}, while substantially reducing the cost of marginal gain estimation.

\noindent \textbf{Sliding-Window Context Approximation.}
Although the full selected subset $\mathcal{S}_{t-1}$ defines the context at step $t$, using all previously selected frames for context-sensitive scoring becomes increasingly expensive as selection proceeds.
Moreover, evaluating each candidate against the entire selection history may impose excessive historical bias, causing later candidates to be overly suppressed.
To alleviate this issue, we maintain a bounded active context window
\begin{equation}
\mathcal{C}_{t-1}\subseteq\mathcal{S}_{t-1},
\qquad |\mathcal{C}_{t-1}| \le W,
\label{eq:context_window}
\end{equation}
where $W$ denotes the window size.
In practice, $\mathcal{C}_{t-1}$ is used as a tractable approximation of the current selection context when computing context-dependent utility terms at step $t$.
Accordingly, both coverage gain and redundancy penalty are evaluated with respect to this active context window during greedy selection.
\begin{table*}[t]
\centering
\caption{
Overview of the six video understanding benchmarks used in our experiments.
}
\label{tab:dataset_overview}
\resizebox{0.95\textwidth}{!}{
\begin{tabular}{lccccc}
\toprule
\textbf{Benchmark} &
\textbf{Video Type} &
\textbf{Dataset Scale} &
\textbf{\# Eval. Samples} &
\textbf{Answer Format} &
\textbf{Metric} \\
\midrule

LongVideoBench
& Multi-domain
& 3,763 videos
& 1,337
& MCQ
& Acc. \\

Video-MME
& Multi-domain
& 900 videos
& 2,700
& MCQ
& Acc. \\

NExT-QA
& Human activities
& 5,440 videos
& 8,576
& MCQ
& Acc. \\

EgoLifeQA
& Egocentric daily life
& $\sim$300 hours
& 3,000
& MCQ
& Acc. \\

YouCook2
& Instructional cooking
& 2,000 videos
& 3,492
& Open-ended
& CIDEr \\

Video-TT
& YouTube Shorts
& 1,000 videos
& 1,000
& Open-ended
& GPT Eval. \\

\bottomrule
\end{tabular}
}
\vspace{-1em}
\end{table*}

\noindent \textbf{Greedy Selection.}
Based on the surrogate marginal utility defined in Eq.~\eqref{eq:overall_utility}, we adopt a greedy strategy that constructs the selected subset incrementally. Figure \ref{fig:main} provides an overview of this greedy selection procedure.
Starting from an empty set $\mathcal{S}_0=\emptyset$, at each step $t$, we evaluate every remaining candidate using the contextual utility in Eq.~\eqref{eq:overall_utility} and select the one with the highest score:
\begin{equation}
f_{i_t}^{*}
=
\arg\max_{f_i\in\mathcal{V}\setminus\mathcal{S}_{t-1}}
u^{(t)}(f_i).
\label{eq:greedy_selection}
\end{equation}
The selected subset is then updated as
\begin{equation}
\mathcal{S}_t
=
\mathcal{S}_{t-1}\cup\{f_{i_t}^{*}\},
\qquad t=1,\ldots,K.
\label{eq:greedy_update}
\end{equation}
This process continues until the frame budget $K$ is reached.
At a high level, each greedy step selects the candidate with the largest estimated marginal contribution under the current context, thereby progressively constructing a subset that is query-relevant, semantically complete, and compact.
The overall procedure is summarized in Alg. \ref{alg:selection}.
\subsection{Discussion}
Let $N$ denote the number of candidate frames, $K$ the target frame budget, $A$ the number of representative anchors, and $W$ the context-window size. 
Computing the coverage gain for all candidates over the entire video requires $\mathcal{O}(N^2)$ computation at each greedy step, leading to $\mathcal{O}(KN^2)$ over $K$ steps. 
Similarly, redundancy computation against all previously selected frames accumulates to $\mathcal{O}(NK^2)$.
MarKey reduces these costs by approximating video-wide coverage with $A$ representative anchors and restricting the context to at most $W$ selected frames, where $A \ll N$ and $W < K$. 
As a result, the coverage and redundancy costs are reduced to $\mathcal{O}(KNA)$ and $\mathcal{O}(KNW)$, respectively.
Therefore, the overall selection complexity is
\begin{equation}
\mathcal{O}\!\left(KN(A+W)\right).
\end{equation}
For fixed $K$, $A$, and $W$, this complexity scales linearly with the number of candidate frames $N$, making iterative subset-aware selection practical for long videos.

\section{Experiments}
\subsection{Experimental Settings}
\noindent\textbf{Datasets.}
We evaluate MarKey on six benchmarks covering holistic video understanding, human-centric video understanding, and open-ended video understanding.
The holistic video understanding benchmarks, LongVideoBench and Video-MME, evaluate broad understanding of diverse long-form video content.
The human-centric benchmarks, NExT-QA and EgoLifeQA, focus on causal-temporal reasoning over human activities and long-horizon understanding of egocentric daily-life videos.
The open-ended benchmarks, YouCook2 and Video-TT, require models to generate responses in natural language, evaluating the generalizability of MarKey beyond predefined answer choices.
Table~\ref{tab:dataset_overview} summarizes the key statistics and evaluation settings of these six benchmarks.

\begin{itemize}

    \item \textit{LongVideoBench}~\citep{wu2024LongVideoBench}
    is a long-context video question-answering benchmark containing videos with temporally aligned subtitles and questions across 17 fine-grained categories.
    In our experiments, we use its validation split, which consists of 1,337 multiple-choice questions involving referred-context reasoning and fine-grained information distributed throughout long video sequences.

    \item \textit{Video-MME}~\citep{fu2025video}
    provides a broad evaluation of video understanding with 900 videos totaling 254 hours and 2,700 expert-annotated multiple-choice question-answer pairs.
    The videos cover six major domains and 30 subfields, with durations ranging from 11 seconds to one hour, and are further divided into short-, medium-, and long-video subsets.

    \item \textit{EgoLifeQA}~\citep{yang2025egolife}
    is constructed from approximately 300 hours of continuous daily-life recordings collected from six participants living together for one week.
    It provides 3,000 long-context multiple-choice questions covering entities, events, habits, interpersonal relationships, and multi-step tasks across extended first-person video histories.

    \item \textit{NExT-QA}~\citep{xiao2021next}
    contains 5,440 videos of daily human activities and 47,692 multiple-choice questions.
    The questions cover causal, temporal, and descriptive understanding, requiring models to identify action dependencies, event order, and relevant interactions among objects and people.

    \item \textit{YouCook2}~\citep{zhou2018towards}
    is a large-scale instructional video benchmark containing 2,000 long, untrimmed videos from 89 cooking recipes.
    In our experiments, we use its open-ended setting, in which models generate natural-language descriptions of cooking procedures.

    \item \textit{Video-TT}~\citep{zhang2025towards}
    consists of 1,000 short-form YouTube videos designed for open-ended video understanding.
    Each video is paired with one open-ended question and four adversarial questions, and we use its open-ended portion in our experiments.

\end{itemize}
\begin{table}[!t]
\centering
\caption{
Comparison of different approaches on holistic video understanding benchmarks.
LongVideoBench (LVB) and Video-MME (V-MME) are evaluated using accuracy (\%).
\textbf{AVG} denotes the average accuracy across the two benchmarks.
All Qwen3-VL-8B-based methods are evaluated using 32 input frames.
Bold numbers indicate the best results among the Qwen3-VL-8B-based methods.
}
\label{tab:comparison_comprehensive}

\renewcommand{\arraystretch}{1.05}
\setlength{\tabcolsep}{4pt}

\begin{tabular*}{\columnwidth}{
    @{\extracolsep{\fill}}
    l
    c
    c
    c
    @{}
}
\toprule
\textbf{Method}
& \textbf{LVB}
& \textbf{V-MME}
& \textbf{AVG} \\
\midrule

\multicolumn{4}{@{}l}{\textit{Reference MLLMs}} \\[-1pt]

GPT-4o~\cite{openai2024gpt4o}
& 66.7 & 71.9 & 69.3 \\

Gemini-1.5-Pro~\cite{team2023gemini}
& 64.0 & 75.0 & 69.5 \\

LLaVA-Video~\cite{zhang2024llava}
& 63.9 & 70.6 & 67.25 \\

LLaVA-OneVision~\cite{li2024llava}
& 59.8 & 68.7 & 64.25 \\

Qwen3-VL-235B-A22B~\cite{bai2025qwen3}
& 65.6 & 79.0 & 72.30 \\

\midrule

\multicolumn{4}{@{}l}{
    \textit{Training-free frame selection with Qwen3-VL-8B}
} \\[-1pt]

Qwen3-VL
& 60.40 & 64.30 & 62.35 \\

Qwen3-VL w/ Top-K
& 62.15 & 64.41 & 63.28 \\

Qwen3-VL w/ AKS~\cite{tang2025adaptive}
& 61.18 & 65.74 & 63.46 \\

Qwen3-VL w/ BOLT~\cite{liu2025bolt}
& 62.52 & 65.37 & 63.95 \\

Qwen3-VL w/ FOCUS~\cite{zhu2025focus}
& 63.05 & 63.15 & 63.10 \\

Qwen3-VL w/ DIG~\cite{li2026divide}
& 64.84 & 65.44 & 65.14 \\

Qwen3-VL w/ AIR~\cite{zou2026air}
& 65.14 & 67.81 & 66.47 \\

\textbf{Qwen3-VL w/ MarKey}
& \textbf{65.74}
& \textbf{68.14}
& \textbf{66.94} \\

\bottomrule
\end{tabular*}

\vspace{-0.8em}
\end{table}

\noindent  \textbf{Evaluated MLLMs and Metrics.}
We use Accuracy (\%) as the primary evaluation metric for multiple-choice video question answering.
For subjective open-ended benchmarks, we follow the original evaluation protocols of each dataset.
Specifically, we use CIDEr as the evaluation metric on YouCook2, and adopt GPT-based evaluation on Video-TT, where the generated response is compared with the ground-truth textual answer to assess semantic correctness.
Following prior studies \citep{team2026script,romero2023zelda}, we further employ Mean Pairwise Cosine Similarity (MPCS) to quantify the similarity among the selected frames. MPCS computes the average cosine similarity over all frame pairs within the selected set, where a lower value indicates less redundancy and greater diversity among the selected frames.
Following prior studies on training-free keyframe selection, we primarily evaluate our method on the Qwen2.5-VL \citep{bai2025qwen2}, Qwen3-VL \citep{bai2025qwen3}, and MiniCPM-V-4\_5 \citep{yu2025minicpm} series, covering different backbone families and model scales.
To provide a broader view of performance, we also report results for a wider range of recent MLLMs on six different long video understanding datasets, including LLaVA-Video \citep{zhang2024llava}, LLaVA-OneVision \citep{li2024llava}, GPT-4o \citep{openai2024gpt4o}, and Gemini \citep{team2023gemini}. 
\begin{table}[!t]
\centering

\caption{
Comparison of different approaches on human-centric video understanding benchmarks.
EgoLifeQA and NExT-QA are evaluated using accuracy (\%).
\textbf{AVG} denotes the average accuracy across the two benchmarks.
All Qwen3-VL-8B-based methods are evaluated using 32 input frames.
Bold numbers indicate the best results among the Qwen3-VL-8B-based methods.
}
\label{tab:comparison_humancentric}

\setlength{\tabcolsep}{2.5pt}
\renewcommand{\arraystretch}{1.08}

\begin{tabular}{@{}lccc@{}}
\toprule
\textbf{Method}
& \textbf{EgoLifeQA}
& \textbf{NExT-QA}
& \textbf{AVG} \\
\midrule

\multicolumn{4}{@{}l}{\textit{Reference MLLMs}} \\[-1pt]

GPT-4o~\cite{openai2024gpt4o}
& 36.2
& --
& -- \\

Gemini-1.5-Pro~\cite{team2023gemini}
& 36.9
& 85.3
& 61.10 \\

LLaVA-Video~\cite{zhang2024llava}
& --
& 85.4
& -- \\

LLaVA-OneVision~\cite{li2024llava}
& --
& 83.2
& -- \\

Qwen3-VL-235B-A22B~\cite{bai2025qwen3}
& --
& 83.3
& -- \\

\midrule

\multicolumn{4}{@{}l}{
\textit{Training-free frame selection with Qwen3-VL-8B}
} \\[-1pt]

Qwen3-VL
& 32.67
& 82.86
& 57.77 \\

Qwen3-VL w/ Top-K
& 32.67
& 82.89
& 57.78 \\

Qwen3-VL w/ AKS~\cite{tang2025adaptive}
& 30.69
& 82.99
& 56.84 \\

Qwen3-VL w/ BOLT~\cite{liu2025bolt}
& 33.66
& 83.08
& 58.37 \\

Qwen3-VL w/ FOCUS~\cite{zhu2025focus}
& 33.66
& 81.39
& 57.53 \\

Qwen3-VL w/ DIG~\cite{li2026divide}
& 33.71
& \textbf{84.61}
& 59.16 \\

Qwen3-VL w/ AIR~\cite{zou2026air}
& 34.65
& 84.31
& 59.48 \\

\textbf{Qwen3-VL w/ MarKey}
& \textbf{37.62}
& 84.20
& \textbf{60.91} \\

\bottomrule
\end{tabular}

\vspace{-1.2em}
\end{table}

\noindent  \textbf{Compared Methods.}
We compare MarKey with several recent training-free keyframe selection methods, including AKS \citep{tang2025adaptive}, BOLT \citep{liu2025bolt}, FOCUS \citep{zhu2025focus}, AIR \citep{zou2026air}, and DIG \citep{li2026divide}.

\begin{itemize}
    \item \textit{AKS} \citep{tang2025adaptive} performs adaptive keyframe selection by recursively partitioning a video according to frame-query similarity. Based on the relevance distribution within different temporal intervals, it dynamically determines the sampling granularity and selects representative frames from informative regions.

    \item \textit{BOLT} \citep{liu2025bolt} adopts a training-free query-guided sampling strategy based on frame-query similarity. It transforms the relevance scores into a probability distribution and samples keyframes accordingly, enabling the selection process to focus more on frames associated with the input query.

    \item \textit{FOCUS} \citep{zhu2025focus} formulates frame selection as a budgeted exploration problem. It progressively evaluates temporal regions and allocates additional sampling resources to promising segments, allowing informative evidence to be identified without densely processing the entire video.

    \item \textit{AIR} \citep{zou2026air} introduces an adaptive iterative reasoning framework for video frame selection. It first identifies potentially relevant temporal regions and then employs multimodal reasoning to iteratively refine the candidate frames, progressively narrowing the search toward useful visual evidence.

    \item \textit{DIG} \citep{li2026divide} adapts the frame selection process according to the characteristics of the input query. It dynamically adjusts the sampling strategy to identify informative visual evidence under different query conditions, providing a flexible training-free solution for long-video understanding.
\end{itemize}

\begin{table}[!t]
\centering

\caption{
Comparison of different approaches on open-ended video understanding benchmarks.
CIDEr is reported for YouCook2, while the GPT-based evaluation score is reported for Video-TT.
\textbf{AVG} denotes the average score across the two benchmarks.
All Qwen3-VL-8B-based methods are evaluated using 32 input frames.
Bold numbers indicate the best results among the Qwen3-VL-8B-based methods.
}
\label{tab:comparison_openended}

\renewcommand{\arraystretch}{1.05}
\setlength{\tabcolsep}{2.5pt}

\begin{tabular*}{0.999\columnwidth}{
    @{\extracolsep{\fill}}
    l
    c
    c
    c
    @{}
}
\toprule
\textbf{Method}
& \textbf{YouCook2}
& \textbf{Video-TT}
& \textbf{AVG} \\
\midrule

\multicolumn{4}{@{}l}{\textit{Reference MLLMs}} \\[-1pt]

GPT-4o~\cite{openai2024gpt4o}
& --
& 36.6
& -- \\

Gemini-1.5-Pro~\cite{team2023gemini}
& 70.4
& 28.8
& 49.60 \\

LLaVA-Video~\cite{zhang2024llava}
& --
& 24.4
& -- \\

\midrule

\multicolumn{4}{@{}l}{
\textit{Training-free frame selection (Qwen3-VL-8B)}
} \\[-1pt]

Qwen3-VL
& 42.02
& 24.10
& 33.06 \\

Qwen3-VL w/ Top-K
& 44.95
& 24.70
& 34.83 \\

Qwen3-VL w/ AKS~\cite{tang2025adaptive}
& 46.45
& 25.10
& 35.78 \\

Qwen3-VL w/ BOLT~\cite{liu2025bolt}
& 45.10
& 27.40
& 36.25 \\

Qwen3-VL w/ FOCUS~\cite{zhu2025focus}
& 43.45
& 26.20
& 34.83 \\

Qwen3-VL w/ DIG~\cite{li2026divide}
& 46.88
& \textbf{32.10}
& \textbf{39.49} \\

Qwen3-VL w/ AIR~\cite{zou2026air}
& 46.37
& 26.50
& 36.43 \\

\textbf{Qwen3-VL w/ MarKey}
& \textbf{48.90}
& 29.70
& 39.30 \\

\bottomrule
\end{tabular*}

\vspace{-0.8em}
\end{table}

\noindent  \textbf{Implementation Details.}
We use the pretrained CLIP-L/14 \citep{radford2021learning} model to extract both visual embeddings for video frames and textual embeddings for input queries. For the Top-K baseline, we rank the same candidate frames as MarKey by their CLIP-L/14 query-frame similarity and select the highest-scoring K frames. For evaluation, we utilize the LMMS-Eval library \citep{zhang2025lmms} to report accuracy on multiple-choice video question answering benchmarks. We set the number of anchors to 648 for anchor-based coverage estimation. We set $\alpha=0.7$, $\delta=0.2$, and $\lambda=0.1$ in all experiments. We set the decoding temperature to 0 for all evaluations.
All experiments are conducted with frame budgets $K \in \{ 16, 32,64,128\}$ on NVIDIA RTX A6000 GPUs with 48 GB of memory.
\begin{table}[!t]
\centering

\caption{
Generalization of different frame selection methods across multiple MLLM backbones on LongVideoBench and Video-MME.
We evaluate the methods with Qwen2.5-VL-7B and MiniCPM-V-4.5-8B to examine whether their effectiveness can consistently transfer across different model architectures.
All frame selection methods within each backbone group are evaluated using 32 input frames.
\textbf{AVG} denotes the average accuracy across LVB and V-MME.
Bold numbers indicate the best results within each backbone.
}
\label{tab:backbone_generalization}

\renewcommand{\arraystretch}{1.05}
\setlength{\tabcolsep}{4pt}

\begin{tabular*}{\columnwidth}{
    @{\extracolsep{\fill}}
    l
    c
    c
    c
    @{}
}
\toprule
\textbf{Method}
& \textbf{LVB}
& \textbf{V-MME}
& \textbf{AVG} \\
\midrule

\multicolumn{4}{@{}l}{\textit{Reference MLLMs}} \\[-1pt]

Gemini-1.5-Pro~\cite{team2023gemini}
& 64.0
& 75.0
& 69.50 \\

InternVL3.5-241B-A28B~\cite{wang2025internvl3}
& 67.1
& 72.9
& 70.00 \\

\midrule

Qwen2.5-VL~\cite{bai2025qwen2}
& 59.31
& 62.20
& 60.76 \\

Qwen2.5-VL w/ AKS~\cite{tang2025adaptive}
& 59.31
& 63.51
& 61.41 \\

Qwen2.5-VL w/ BOLT~\cite{liu2025bolt}
& 60.00
& 63.67
& 61.84 \\

Qwen2.5-VL w/ FOCUS~\cite{zhu2025focus}
& 61.48
& 62.37
& 61.93 \\

\textbf{Qwen2.5-VL w/ MarKey}
& \textbf{63.94}
& \textbf{65.03}
& \textbf{64.49} \\

\midrule

MiniCPM-V-4.5
& 60.61
& 63.88
& 62.25 \\

MiniCPM-V-4.5 w/ AKS~\cite{tang2025adaptive}
& 60.81
& 64.81
& 62.81 \\

MiniCPM-V-4.5 w/ BOLT~\cite{liu2025bolt}
& 60.94
& 64.40
& 62.67 \\

MiniCPM-V-4.5 w/ FOCUS~\cite{zhu2025focus}
& 61.56
& 64.11
& 62.84 \\

\textbf{MiniCPM-V-4.5 w/ MarKey}
& \textbf{64.15}
& \textbf{66.33}
& \textbf{65.24} \\

\midrule

\textbf{Qwen3-VL w/ MarKey}
& \textbf{65.74}
& \textbf{68.14}
& \textbf{66.94} \\

\bottomrule
\end{tabular*}

\vspace{-0.8em}
\end{table}

\subsection{Main Results}
To comprehensively evaluate the effectiveness and generalizability of MarKey, we compare it with representative training-free frame selection methods, including Top-K, AKS, BOLT, FOCUS, DIG, and AIR, under the same Qwen3-VL-8B backbone and a fixed input budget of 32 frames. For a more structured comparison, we organize the six benchmarks into three complementary categories: holistic video understanding, specialized human-centric video understanding, and open-ended video understanding.

\paragraph{Holistic Video Understanding.}
As shown in Table~\ref{tab:comparison_comprehensive}, MarKey achieves the best performance among all Qwen3-VL-8B-based methods on both LongVideoBench and Video-MME, with an average accuracy of 66.94\%.
First, MarKey outperforms Top-K by 3.59 and 3.73 points on LongVideoBench and Video-MME, respectively. Since Top-K and MarKey use the same CLIP-based query relevance, this comparison directly shows that frame-wise relevance alone is insufficient. Accounting for the complementarity among selected frames leads to consistently better use of the same frame budget. 
Second, the relative performance of existing selectors varies across the two benchmarks. For example, FOCUS is more competitive on LongVideoBench, while AKS shows a stronger relative advantage on Video-MME. MarKey achieves the best performance on both benchmarks, indicating that its effectiveness is less dependent on a particular benchmark or evidence distribution.
Third, MarKey substantially improves the competitiveness of a relatively small backbone. On LongVideoBench, Qwen3-VL-8B with MarKey reaches 65.74\%, slightly surpassing Qwen3-VL-235B-A22B at 65.60\% under the same 32-frame budget, despite the nearly 30$\times$ difference in model size. This result highlights the importance of input evidence quality in long-video understanding. Under a constrained visual budget, allocating frames more effectively can yield gains comparable to those obtained by substantially increasing model capacity, making frame selection an effective way to improve performance without scaling the downstream MLLM.
\begin{figure}[t]
\centering
\includegraphics[width=0.478\textwidth, trim=0 0 0 0]{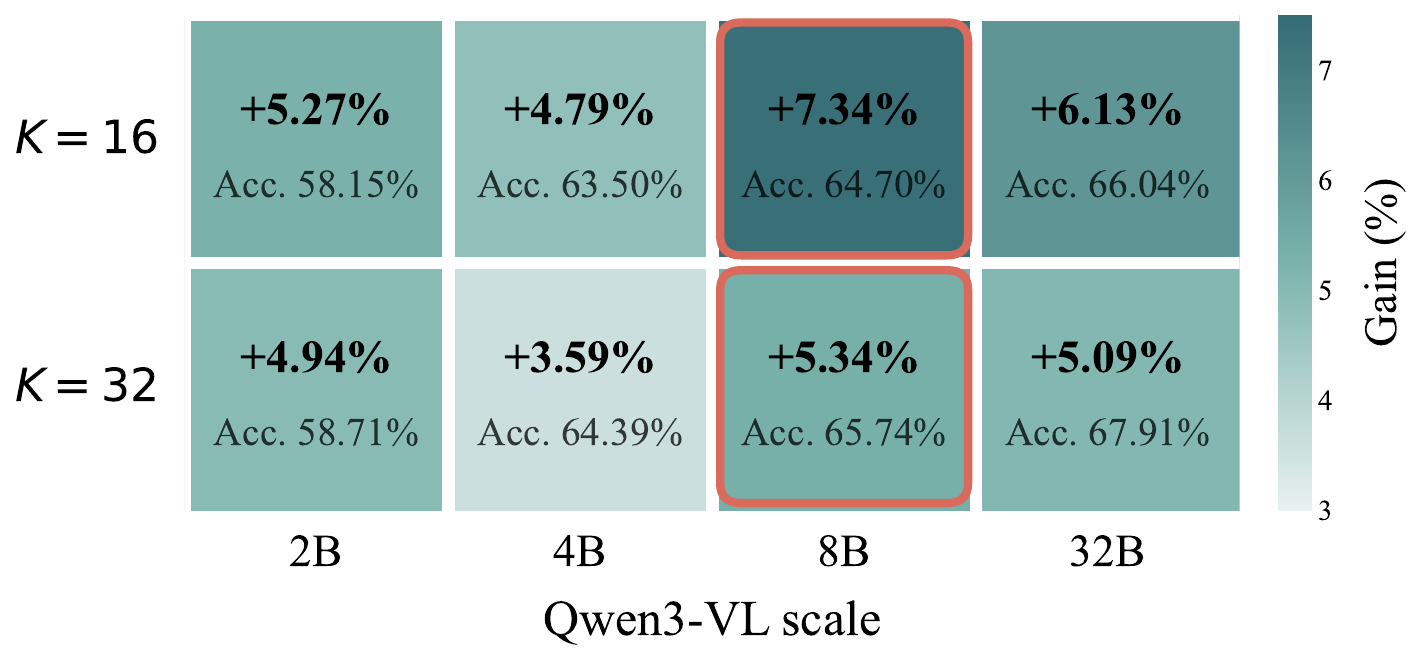}
\caption{Comparison of different LVLM scales using Qwen3-VL on LongVideoBench. Accuracy (\%) is reported. Results show performance with $K$=16 frames (top) and $K$=32 frames (bottom). }
\label{figure:weight}
\vspace{-2em}
\end{figure}

\paragraph{Human-centric Video Understanding.}
As shown in Table~\ref{tab:comparison_humancentric}, MarKey achieves the highest average accuracy of 60.91\% among the Qwen3-VL-8B-based methods, demonstrating strong performance across two substantially different human-centric video understanding settings.
First, the improvement is particularly pronounced on EgoLifeQA. MarKey improves the Qwen3-VL baseline from 32.67\% to 37.62\%, yielding a gain of 4.95 points, and exceeds the strongest competing selector, AIR, by 2.97 points. EgoLifeQA requires reasoning over extended first-person video histories, where relevant evidence may occur at distant moments and multiple observations may need to be combined. The larger gain on this benchmark therefore supports the importance of preserving complementary evidence under a limited frame budget.
Second, MarKey achieves strong overall performance across both human-centric benchmarks, while the relative improvements differ between them. This result reflects the different characteristics and challenges of the two benchmarks, and further demonstrates the robustness of MarKey across diverse human-centric video understanding scenarios.

\paragraph{Open-ended Video Understanding.}
Table~\ref{tab:comparison_openended} further evaluates whether the benefits of MarKey extend beyond multiple-choice question answering to settings that require models to generate natural-language responses.
First, MarKey consistently improves the Qwen3-VL baseline on both open-ended benchmarks, increasing the YouCook2 CIDEr score from 42.02 to 48.90 and the Video-TT score from 24.10 to 29.30. The gains of 6.88 and 5.20 points, respectively, show that selecting more informative and complementary frames also benefits free-form generation, where the model cannot rely on predefined answer candidates.
Second, MarKey achieves the best performance on YouCook2, outperforming the strongest competing method, DIG, by 2.02 points. This result is particularly relevant for instructional videos, where generating an accurate description requires sufficient coverage of multiple procedural steps rather than identifying only a single salient moment. The improvement therefore aligns well with MarKey's explicit modeling of coverage gain and redundancy.
Third, MarKey shows a larger advantage on YouCook2 than on Video-TT. Relative to Video-TT, YouCook2 contains longer and more procedurally structured videos, where relevant evidence can be distributed across multiple stages. This result further highlights the strength of MarKey in preserving complementary information across extended video content.
\subsection{Effect of Different Backbones}
To examine whether the effectiveness of MarKey depends on a specific MLLM backbone, we further evaluate it using Qwen2.5-VL-7B and MiniCPM-V-4.5-8B on LongVideoBench and Video-MME. All methods use the same input budget of 32 frames, and the results are reported in Table \ref{tab:backbone_generalization}.
First, MarKey consistently achieves the best performance across both backbones and benchmarks. Compared with uniform sampling, it improves Qwen2.5-VL by 4.63 points on LongVideoBench and 2.83 points on Video-MME, while improving MiniCPM-V-4.5 by 3.54 and 2.45 points, respectively.
Second, compared with the strongest competing frame selector, MarKey further improves performance by 2.46 points on LongVideoBench and 1.36 points on Video-MME under Qwen2.5-VL, and by 2.59 and 1.52 points under MiniCPM-V-4.5, respectively. These results show that the effectiveness of MarKey generalizes across different MLLM backbones rather than being specific to Qwen3-VL.
\begin{figure}[t]
\centering
\includegraphics[width=0.478\textwidth, trim=0 0 0 0]{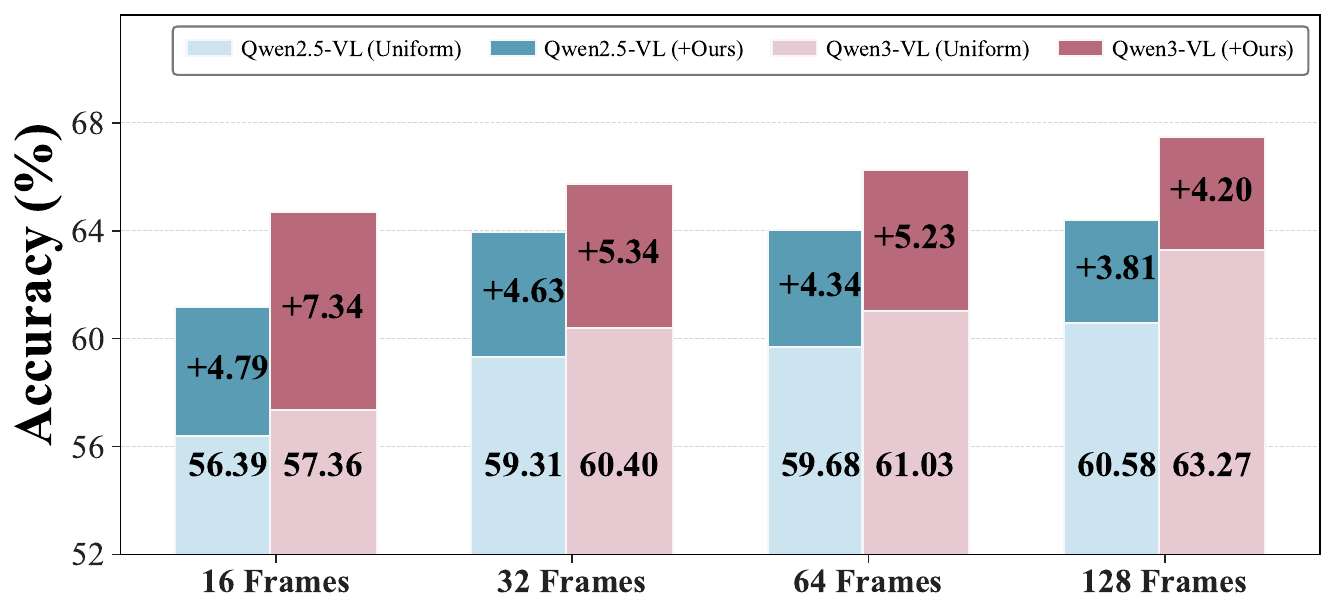}
\caption{Performance comparison across different frame budgets K on LongVideoBench. MarKey consistently outperforms uniform sampling across all LVLMs and budget settings.}
\label{fig:keyframe}
\vspace{-1em}
\end{figure}
Third, we observe consistently larger improvements on LongVideoBench than on Video-MME. This suggests that subset-aware selection is particularly beneficial when useful evidence is distributed over longer temporal contexts, where effective evidence allocation becomes more important under a fixed frame budget.
\begin{table}[t]
\centering
\caption{Component ablation study on LongVideoBench and Video-MME
with Qwen3-VL-8B ($K{=}32$). All scores are accuracy (\%).}
\label{tab:ablation}

\setlength{\tabcolsep}{2.5pt}
\renewcommand{\arraystretch}{1.05}

\begin{tabular}{@{}ccc|cc@{}}
\toprule
\multicolumn{3}{c|}{\textbf{Components}}
& \multirow{2}{*}{\textbf{LVB}}
& \multirow{2}{*}{\textbf{V-MME}} \\
\cmidrule(lr){1-3}
\textbf{Query} & \textbf{Coverage} & \textbf{Redundancy}
& & \\
\midrule
$\times$     & $\times$     & $\times$     & 60.40 & 64.30 \\
\checkmark   & $\times$     & $\times$     & 62.15 & 64.41 \\
$\times$     & \checkmark   & $\times$     & 60.87 & 65.11 \\
\checkmark   & \checkmark   & $\times$     & 64.17 & 65.40 \\
\checkmark   & $\times$     & \checkmark   & 64.02 & 66.07 \\
$\times$     & \checkmark   & \checkmark   & 61.01 & 64.81 \\
\checkmark   & \checkmark   & \checkmark
& \textbf{65.74} & \textbf{68.14} \\
\bottomrule
\end{tabular}
\vspace{-1em}
\end{table}

\subsection{Effect of Different Model Scales}
To further examine the scalability of our method, we evaluate it on Qwen3-VL models at different scales, including 2B, 4B, 8B and 32B, and report the results in Figure \ref{figure:weight}. Several observations can be drawn from the results.
First, our method consistently improves performance across all model scales on LongVideoBench, suggesting that its effectiveness is stable and not tied to a specific parameter budget.
Second, we observe an interesting phenomenon under uniform sampling: Qwen3-VL-4B slightly outperforms Qwen3-VL-8B in the baseline setting. However, after replacing uniform sampling with our keyframe selection strategy, the 8B model exhibits the largest improvement among the tested scales.
We conjecture that, under a fixed frame budget, the performance of larger models is more sensitive to the quality of visual inputs. When uniformly sampled frames contain redundant or less informative content, the stronger reasoning capacity of the larger model cannot be fully utilized.
In contrast, once more informative keyframes are provided, the 8B model is better able to exploit its higher capacity for temporal evidence aggregation and question-conditioned reasoning. This suggests that effective frame selection is particularly important for unlocking the potential of larger MLLMs.
\subsection{Effect of Keyframe Number}
To examine the impact of the number of input keyframes, we evaluate our method under different frame budgets, including 16, 32, 64, and 128 frames. The results are shown in Figure \ref{fig:keyframe}. Several observations can be drawn from the results.
First, our method consistently outperforms uniform sampling across all frame budgets on both Qwen2.5-VL and Qwen3-VL, indicating that its effectiveness is robust to different input lengths rather than relying on a specific frame budget. 
Second, our method remains effective even with a large number of input frames. Notably, under the 128-frame setting, our method achieves improvements of 3.81 and 4.20 percentage points over uniform sampling on Qwen2.5-VL-7B and Qwen3-VL-8B, respectively. These results demonstrate that increasing the visual budget alone is insufficient to fully exploit long videos, while selecting informative keyframes remains essential for effective video understanding.
\begin{figure}[t]
\centering
\includegraphics[width=0.478\textwidth, trim=0 0 0 0]{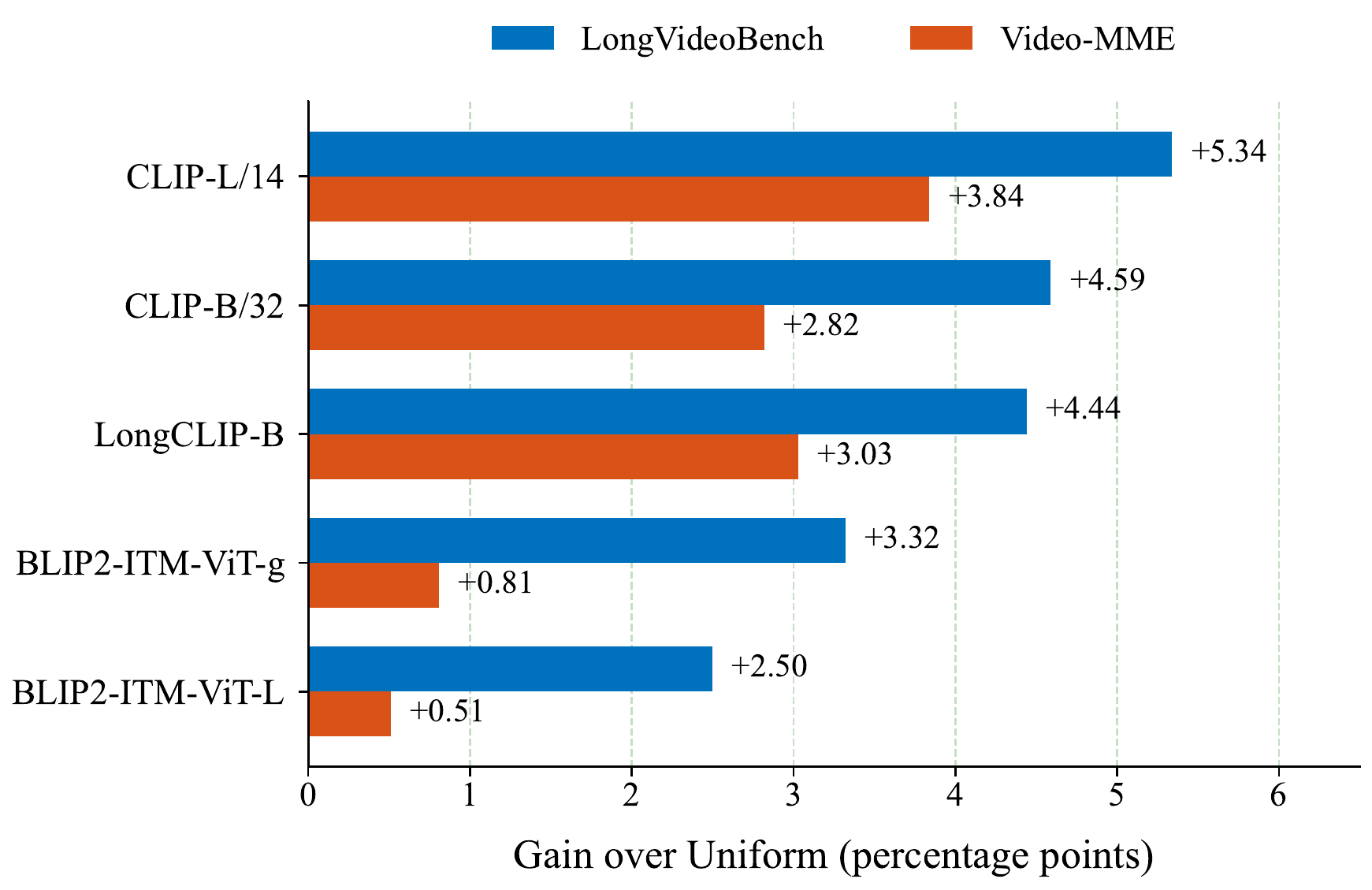}
\caption{Ablation on different VLMs for query-frame similarity scoring on LongVideoBench and Video-MME with Qwen3-VL-8B ($K$=32). Accuracy scores (\%) are reported. CLIP-L/14 achieves the best performance on both Video-MME and LongVideoBench. Therefore, we use CLIP-L/14 as the default scorer.}
\label{figure:vlm_comparison}
\vspace{-1em}
\end{figure}
\subsection{Ablation Study}
To investigate the contribution of each component in our proposed method, we conduct five different ablation variants, including individually removing Query Relevance, Visual Coverage, and Redundancy Penalty, as well as retaining only Query Relevance or Visual Coverage. We further compare these different variants with both the complete model and the uniform sampling baseline under the same experimental setting. The corresponding results are summarized in Table \ref{tab:ablation}. Based on these experimental results, several important observations can be drawn.
First, removing any individual component consistently leads to noticeable performance degradation, and none of the ablated variants can match the full model. This verifies that the performance gain of MarKey does not come from a single dominant design choice, but rather from the effective interaction of all three components.
More importantly, Query Relevance is shown to be the most critical component among the three components. Removing it causes the score to substantially drop from 65.74\% to 61.01\% on LongVideoBench, corresponding to a considerable decrease of 4.73 points and making the resulting performance only marginally better than uniform sampling (60.40\%). This result clearly suggests that query-aware relevance modeling provides the fundamental and essential signal for effective keyframe selection, since it enables the selector to more effectively focus on visual evidence that is directly relevant and useful for accurately answering the given question.
Furthermore, the ablation results highlight the complementary effect of the three components. Using Query Relevance or Visual Coverage alone yields only limited improvements, indicating that either signal in isolation is insufficient for reliable keyframe selection. In contrast, combining relevance with contextual coverage and redundancy modeling consistently leads to stronger performance. This suggests that task-relevant evidence can be more effectively identified when candidate frames are evaluated within a sufficiently informative selection context, where both complementary content and repeated evidence are explicitly considered.

\begin{table}[t]
\centering
\caption{Ablation of utility weights on LongVideoBench and Video-MME with Qwen3-VL-8B ($K{=}32$). Accuracy (\%) is reported. $\alpha$, $\delta$, and $\lambda$ denote the weights for query relevance, visual coverage, and redundancy penalty, respectively.}
\setlength{\tabcolsep}{8pt}
\renewcommand{\arraystretch}{1.15}
\begin{tabular}{c c c c c}
\toprule
$\boldsymbol{\alpha}$ & $\boldsymbol{\delta}$ & $\boldsymbol{\lambda}$ & \textbf{LVB} & \textbf{V-MME} \\
\midrule
\multicolumn{3}{c}{Uniform Sampling} & 60.40 & 64.30 \\
\midrule
0.90 & 0.05 & 0.05 & 64.54& 66.77 \\
0.85 &0.10 & 0.05 & 64.77 & 66.81\\
0.80 & 0.15 & 0.05 & 64.69 & 66.70\\
0.75 &0.15 & 0.10 & 65.07 &67.40 \\
0.60 & 0.20 & 0.20 & 63.94 & 67.25\\
0.55 &0.25 & 0.20 & 64.09 & 66.96\\
\midrule
\textbf{0.70} & \textbf{0.20}& \textbf{0.10} & \textbf{65.74} & \textbf{68.14}  \\
\bottomrule
\end{tabular}
\label{tab:weight_ablation}
\end{table}
\subsection{Effect of Different VLM Scorer}
To evaluate the impact of the VLM scorer used for computing query-frame relevance, we compare several representative vision-language models, including CLIP-B/32 \citep{radford2021learning}, CLIP-L/14 \citep{radford2021learning}, LongCLIP-B \citep{zhang2024long}, BLIP2-ITM-ViT-L \citep{li2023blip}, and BLIP2-ITM-ViT-g \citep{li2023blip}.
The results are summarized in Figure \ref{figure:vlm_comparison}. Several observations can be drawn from the results.
First, the choice of VLM scorer has a clear impact on the final performance, indicating that the quality of query-frame relevance estimation is important for effective frame selection.
Second, all tested VLM scorers consistently outperform uniform sampling on both LongVideoBench and Video-MME, while the magnitude of improvement varies substantially across different scorers. In particular, BLIP2-based scorers provide relatively modest gains, whereas LongCLIP-B and the standard CLIP variants yield considerably larger improvements. This suggests that different pretrained vision-language models exhibit different levels of effectiveness in estimating query-frame relevance, further highlighting the importance of selecting an appropriate scorer for query-aware frame selection.
Third, CLIP-L/14 delivers the best overall results, achieving 65.74\% on LongVideoBench and 68.14\% on Video-MME, while CLIP-B/32 also remains highly competitive. This trend suggests that CLIP-based scorers provide stronger and more reliable cross-modal relevance signals for our method than the BLIP2 variants in this work.
\begin{figure}[t]
\centering
\includegraphics[width=0.478\textwidth, trim=0 0 0 0]{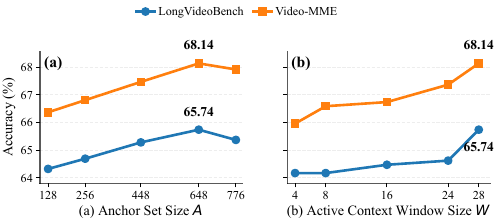}
\caption{Ablation studies of the anchor set size and active context window size on LongVideoBench and Video-MME with Qwen3-VL-8B ($K{=}32$). Accuracy (\%) is reported. (a) Effect of the anchor set size, where $A$ denotes the number of anchors used for coverage estimation. (b) Effect of the active context window size, where $W$ denotes the number of selected frames retained in the active context during greedy selection.}
\label{fig:window}
\end{figure}
\subsection{Analysis of Hyperparameters}
\noindent{\textbf{Utility Coefficients.}}
To explore the sensitivity of MarKey to the weighting coefficients in the utility function, we vary the values of $\alpha$, $\delta$, and $\lambda$, which control query relevance, visual coverage, and redundancy penalty, respectively. The results are reported in Table \ref{tab:weight_ablation}. Several observations can be drawn from the results. First, all tested weight configurations consistently outperform the uniform sampling baseline on both LongVideoBench and Video-MME, demonstrating that the proposed utility design is effective and robust across a reasonable range of coefficient choices.
Second, our default setting, $\alpha{=}0.70$, $\delta{=}0.20$, and $\lambda{=}0.10$, achieves the best overall trade-off across the two benchmarks, obtaining the highest score on LongVideoBench (65.74\%) and Video-MME (68.14\%).
Third, the results highlight that query relevance, visual coverage, and redundancy suppression are complementary to each other in the proposed utility design. Their joint modeling enables MarKey to select subsets that are more query-relevant, more complete, and less redundant, which further supports our formulation of keyframe selection as a context-aware subset selection problem.

\noindent\textbf{Anchor Set Size.}
To investigate the effect of the anchor set size, we vary $A$ while keeping all other settings fixed. As shown in Figure \ref{fig:window}(a), performance improves as $A$ increases from 128 to 648, indicating that a larger anchor set provides a more reliable approximation of global visual coverage. However, further increasing $A$ to 776 slightly reduces the accuracy on both benchmarks.
This result suggests that excessively dense anchors may introduce redundant coverage information without providing additional benefit. We therefore set $A=648$ as the default, which achieves the best performance while avoiding unnecessary computation.

\noindent{\textbf{Window Size.}}
To explore the effect of the active context window size, we vary $W$ while keeping all other settings fixed. The results are reported in Figure \ref{fig:window}(b). Several observations can be drawn from the results. First, the performance improves generally as the window size increases, indicating that a larger active context is beneficial for modeling the interaction between the current candidate and the previously selected frames. On LongVideoBench, the score improves from 64.17\% at $W{=}4$ to 65.74\% at the default setting $W{=}28$. A similar trend is observed on Video-MME, where the score rises from 65.96\% to 68.14\%. These results suggest that incorporating a broader selection context helps estimate the contextual utility of candidate frames more accurately.
Second, enlarging the window size also increases the computational cost of context-dependent scoring. At the same time, the performance improvement becomes less substantial when the window size is already large. This trend suggests that, in practice, using a moderately sized active context is often sufficient to capture most of the useful contextual information for candidate evaluation. In this way, MarKey can maintain strong performance while reducing the cost of context-dependent scoring. Such a design is particularly appealing in more complex scenarios, where selecting a larger number of keyframes would otherwise make full-context evaluation increasingly expensive.

\begin{figure}[t]
\centering
\includegraphics[width=0.498\textwidth, trim=0 0 0 0]{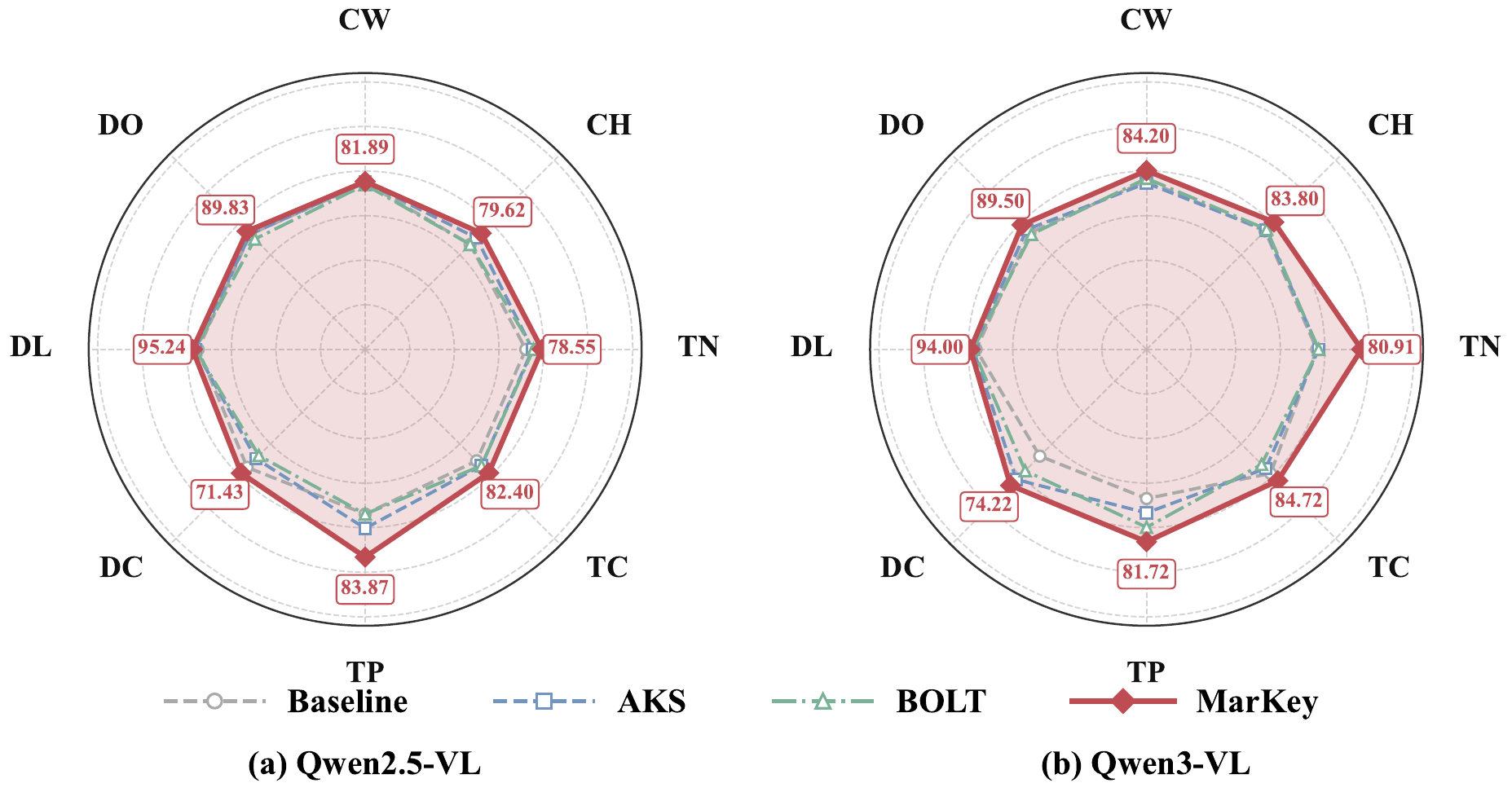}
\caption{
Fine-grained accuracy (\%) comparison across eight question types on NExT-QA using two vision-language models with 32 input frames: (a) Qwen2.5-VL and (b) Qwen3-VL-8B. CW and CH denote Causal Why and Causal How; TN, TC, and TP denote Temporal Next, Temporal Current, and Temporal Previous; DC, DL, and DO denote Descriptive Count, Descriptive Location, and Descriptive Other, respectively.
}
\label{fig:nextqa_type_analysis}
\end{figure}
\subsection{Analysis of Video Duration}
\begin{table}[t]
\centering
\caption{Comparison of different keyframe selection methods on LongVideoBench validation under different video duration groups for Qwen3-VL-8B. Short denotes videos shorter than 3 minutes, Medium denotes videos from 3 to 20 minutes, and Long denotes videos longer than 20 minutes. Accuracy scores (\%) are reported.}
\label{tab:vlm_duration_comparison}
\begin{tabular}{l|ccc}
\toprule
\textbf{Method} & \textbf{Short} & \textbf{Medium} & \textbf{Long} \\
\midrule
Qwen3-VL w/ Uniform & \textbf{74.79} & 59.46 & 51.77 \\
\midrule
Qwen3-VL w/ AKS    & 73.96 & 60.44 & 53.55 \\
Qwen3-VL w/ BOLT   & 74.52 & 63.11 & 54.43 \\
Qwen3-VL w/ FOCUS  & 70.36 & 64.80 & 57.45 \\
Qwen3-VL w/ DIG  & 73.68& 62.85 & 57.97 \\
Qwen3-VL w/ AIR   & 73.96& 65.14 & 53.62 \\
\textbf{Qwen3-VL w/ MarKey} & 74.06 & \textbf{65.77} & \textbf{59.57} \\
\bottomrule
\end{tabular}
\end{table}

\begin{figure*}[t]
\centering
\includegraphics[width=0.99\textwidth, trim=0 0 0 0]{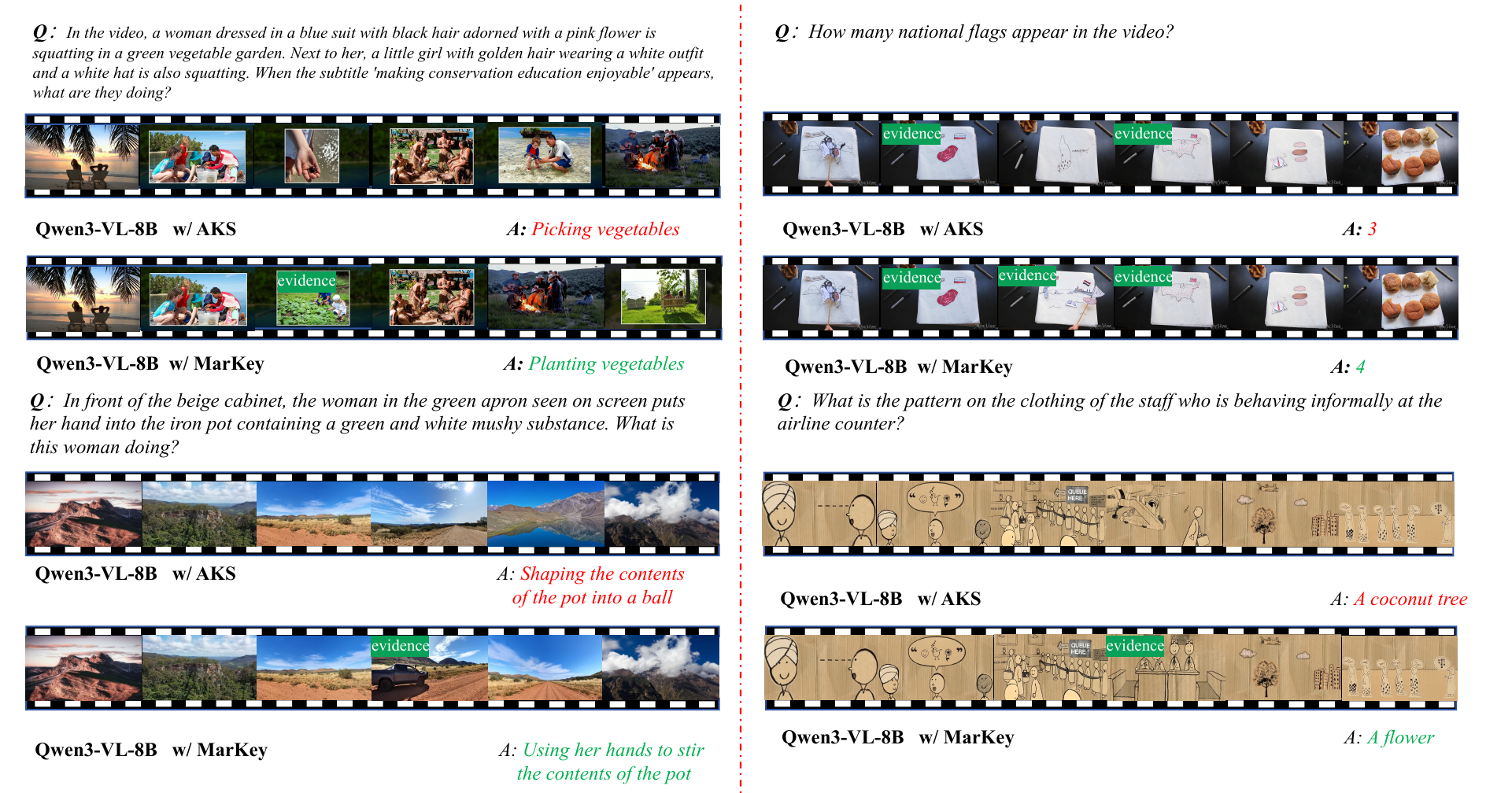}
\caption{
Comparison between frames selected by AKS and MarKey on representative video understanding examples.
The left column shows two examples from LongVideoBench, while the right column shows two from Video-MME. Green boxes indicate the question-relevant evidence frames selected by MarKey. Compared with AKS, MarKey more effectively captures informative evidence relevant to the query, leading to more accurate answers under the same frame budget.
}
\label{fig:case}
\end{figure*}
To examine the effect of video duration on keyframe selection, we compare different training-free methods on LongVideoBench validation under three duration groups, including short, medium, and long videos. The results are shown in Table \ref{tab:vlm_duration_comparison}. Several observations can be drawn from the results.
First, MarKey achieves the best overall performance and delivers the strongest results on medium and long videos, while remaining competitive on short videos.
Second, the advantage of MarKey becomes increasingly pronounced as video duration grows. The improvements on medium and long videos are much more substantial than those on short videos. In particular, compared with uniform sampling, MarKey improves accuracy by 6.31 points on medium videos and 7.80 points on long videos, clearly showing that its benefit becomes more evident as the temporal span increases.
This trend further demonstrates the effectiveness of MarKey for long-video understanding in more challenging settings. As videos become longer, useful evidence is often more temporally dispersed and redundant visual content becomes more prevalent, making frame selection increasingly challenging under a fixed budget. In such cases, the explicit modeling of coverage and redundancy in MarKey helps preserve more complete task-relevant evidence while reducing unnecessary overlap, thereby enabling more effective, robust, and reliable keyframe selection for long-horizon reasoning.
\subsection{Fine-grained Analysis on NExT-QA} 
To further understand the effectiveness of MarKey under different reasoning requirements, we conduct a fine-grained evaluation on the eight question types of NExT-QA, covering causal, temporal, and descriptive reasoning. The results are shown in Figure \ref{fig:nextqa_type_analysis}.
First, MarKey achieves the best performance across all eight question types, indicating that its overall improvement is not dominated by a single category. Consistent gains are observed on causal, temporal, and descriptive questions, demonstrating its robustness across diverse reasoning requirements.
Second, MarKey exhibits particularly pronounced improvements on temporal reasoning questions across both Qwen2.5-VL and Qwen3-VL. For example, compared with the original Qwen3-VL baseline, MarKey improves Temporal Next and Temporal Previous by 3.35 and 3.23 percentage points, respectively, representing the two largest gains among the eight question types. This indicates that MarKey is especially effective at preserving the temporally complementary evidence required for reasoning about preceding and subsequent events.
\begin{table}[t]
\centering
\caption{Comparison of Mean Pairwise Cosine Similarity (MPCS) on LongVideoBench and Video-MME. All methods use Qwen3-VL-8B with 32 selected frames.}
\label{tab:mpcs_comparison}
\begin{tabular}{lcc}
\toprule
\textbf{Method} & \textbf{LVB}  & \textbf{V-MME} \\
\midrule
Qwen3-VL w/ AKS   & 0.6539 & 0.7161 \\
Qwen3-VL w/ BOLT  & 0.7164 & 0.7975 \\
Qwen3-VL w/ FOCUS & 0.7490 & 0.7655 \\
Qwen3-VL w/ DIG &  0.7617 &  0.7332  \\
Qwen3-VL w/ AIR &  0.7457  & 0.7459 \\
\textbf{Qwen3-VL w/ MarKey} & \textbf{0.6177} & \textbf{0.6773} \\
\bottomrule
\end{tabular}
\end{table}

\subsection{Analysis of Selected-Frame Redundancy}
To further examine the redundancy of the selected keyframes, we evaluate the Mean Pairwise Cosine Similarity among the selected frames on LongVideoBench and Video-MME. All methods use Qwen3-VL-8B with a fixed budget of 32 input frames. The results are shown in Table~\ref{tab:mpcs_comparison}. Lower MPCS indicates lower similarity among the selected frames and thus less inter-frame redundancy. Several observations can be drawn from the results.
We can observe that MarKey achieves the lowest MPCS on both benchmarks, with 0.6177 on LongVideoBench and 0.6773 on Video-MME, outperforming the strongest competing method by 0.0362 and 0.0388, respectively. This indicates that MarKey selects less redundant and more complementary frames.
Moreover, MarKey achieves the best downstream performance while maintaining the lowest MPCS, supporting our central claim that reducing redundant selections and preserving complementary evidence leads to more effective use of the limited visual budget.

\subsection{Case Study}
To provide a qualitative comparison, Figure \ref{fig:case} visualizes the keyframes selected by MarKey and AKS on representative examples from LongVideoBench and Video-MME.
In these examples, MarKey tends to retain frames that are more directly and consistently relevant to the question while providing broader coverage of the video content and avoiding visually repetitive selections. As a result, the selected subsets are not only better aligned with the query, but also more likely to capture complementary evidence distributed across different moments of the video.
This advantage is particularly evident in the Video-MME examples, where answering the question often requires aggregating evidence from multiple temporal locations rather than relying on a single salient moment. For example, when asked how many national flags appear in the video, AKS misses part of the relevant evidence and leads the MLLM to predict three, whereas MarKey preserves more complementary evidence across different moments and enables the model to recover the correct count of four. This example highlights the importance of maintaining broad yet query-relevant temporal coverage when the supporting evidence is distributed throughout the video.
Overall, these qualitative examples support the quantitative findings and further illustrate that MarKey improves video understanding not simply by changing which frames are sampled, but by constructing a more effective, query-relevant, informative, less redundant, and more globally representative visual subset for downstream reasoning under a limited frame budget.

\section{Conclusions}
In this paper, we introduce MarKey, a selected subset-aware greedy keyframe selection framework for long-video understanding.
MarKey reformulates keyframe selection as a context-dependent subset valuation problem, where the importance of a frame is measured by its approximate marginal contribution to the currently selected subset rather than by its standalone relevance.
To make this formulation tractable, MarKey adopts a surrogate utility that jointly models query relevance, visual coverage, and redundancy suppression, and further employs temporally sampled anchor frames together with a greedy selection strategy to efficiently construct compact yet informative keyframe subsets under a fixed frame budget.
Extensive experiments across six video-understanding benchmarks demonstrate that MarKey consistently improves the performance of strong long-video MLLMs and outperforms existing training-free keyframe selection baselines.
\section*{Acknowledgments}
This work was supported in part by the National Natural Science Foundation of China under Grant 62402158, and in part by the Key Science \& Technology Project of Anhui Province under Grant 202523j08050001.
{
    \small
    \bibliographystyle{ieeenat_fullname}
    \bibliography{main}

@String(AAAI = {AAAI})

@article{liu2023visual,
  title={Visual instruction tuning},
  author={Liu, Haotian and Li, Chunyuan and Wu, Qingyang and Lee, Yong Jae},
  journal={Advances in neural information processing systems},
  volume={36},
  pages={34892--34916},
  year={2023}
}

@article{zhu2023minigpt,
  title={Minigpt-4: Enhancing vision-language understanding with advanced large language models},
  author={Zhu, Deyao and Chen, Jun and Shen, Xiaoqian and Li, Xiang and Elhoseiny, Mohamed},
  journal={arXiv preprint arXiv:2304.10592},
  year={2023}
}

@article{li2025videochat,
  title={Videochat: Chat-centric video understanding},
  author={Li, KunChang and He, Yinan and Wang, Yi and Li, Yizhuo and Wang, Wenhai and Luo, Ping and Wang, Yali and Wang, Limin and Qiao, Yu},
  journal={Science China Information Sciences},
  volume={68},
  number={10},
  pages={200102},
  year={2025},
  publisher={Springer}
}

@inproceedings{maaz2024video,
  title={Video-chatgpt: Towards detailed video understanding via large vision and language models},
  author={Maaz, Muhammad and Rasheed, Hanoona and Khan, Salman and Khan, Fahad},
  booktitle={Proceedings of the 62nd Annual Meeting of the Association for Computational Linguistics (Volume 1: Long Papers)},
  pages={12585--12602},
  year={2024}
}

@inproceedings{lin2024video,
  title={Video-llava: Learning united visual representation by alignment before projection},
  author={Lin, Bin and Ye, Yang and Zhu, Bin and Cui, Jiaxi and Ning, Munan and Jin, Peng and Yuan, Li},
  booktitle={Proceedings of the 2024 conference on empirical methods in natural language processing},
  pages={5971--5984},
  year={2024}
}

@inproceedings{zhang2023video,
  title={Video-llama: An instruction-tuned audio-visual language model for video understanding},
  author={Zhang, Hang and Li, Xin and Bing, Lidong},
  booktitle={Proceedings of the 2023 conference on empirical methods in natural language processing: system demonstrations},
  pages={543--553},
  year={2023}
}

@article{li2024llava,
  title={Llava-onevision: Easy visual task transfer},
  author={Li, Bo and Zhang, Yuanhan and Guo, Dong and Zhang, Renrui and Li, Feng and Zhang, Hao and Zhang, Kaichen and Zhang, Peiyuan and Li, Yanwei and Liu, Ziwei and others},
  journal={arXiv preprint arXiv:2408.03326},
  year={2024}
}

@article{li2024llava1,
  title={Llava-next-interleave: Tackling multi-image, video, and 3d in large multimodal models},
  author={Li, Feng and Zhang, Renrui and Zhang, Hao and Zhang, Yuanhan and Li, Bo and Li, Wei and Ma, Zejun and Li, Chunyuan},
  journal={arXiv preprint arXiv:2407.07895},
  year={2024}
}

@article{li2024aria,
  title={Aria: An open multimodal native mixture-of-experts model},
  author={Li, Dongxu and Liu, Yudong and Wu, Haoning and Wang, Yue and Shen, Zhiqi and Qu, Bowen and Niu, Xinyao and Zhou, Fan and Huang, Chengen and Li, Yanpeng and others},
  journal={arXiv preprint arXiv:2410.05993},
  year={2024}
}

@article{xu2024pllava,
  title={Pllava: Parameter-free llava extension from images to videos for video dense captioning},
  author={Xu, Lin and Zhao, Yilin and Zhou, Daquan and Lin, Zhijie and Ng, See Kiong and Feng, Jiashi},
  journal={arXiv preprint arXiv:2404.16994},
  year={2024}
}

@article{liu2024kangaroo,
  title={Kangaroo: A powerful video-language model supporting long-context video input},
  author={Liu, Jiajun and Wang, Yibing and Ma, Hanghang and Wu, Xiaoping and Ma, Xiaoqi and Wei, Xiaoming and Jiao, Jianbin and Wu, Enhua and Hu, Jie},
  journal={arXiv preprint arXiv:2408.15542},
  year={2024}
}

@article{chen2024longvila,
  title={Longvila: Scaling long-context visual language models for long videos},
  author={Chen, Yukang and Xue, Fuzhao and Li, Dacheng and Hu, Qinghao and Zhu, Ligeng and Li, Xiuyu and Fang, Yunhao and Tang, Haotian and Yang, Shang and Liu, Zhijian and others},
  journal={arXiv preprint arXiv:2408.10188},
  year={2024}
}

@article{zhang2024long,
  title={Long context transfer from language to vision},
  author={Zhang, Peiyuan and Zhang, Kaichen and Li, Bo and Zeng, Guangtao and Yang, Jingkang and Zhang, Yuanhan and Wang, Ziyue and Tan, Haoran and Li, Chunyuan and Liu, Ziwei},
  journal={arXiv preprint arXiv:2406.16852},
  year={2024}
}

@inproceedings{weng2024longvlm,
  title={Longvlm: Efficient long video understanding via large language models},
  author={Weng, Yuetian and Han, Mingfei and He, Haoyu and Chang, Xiaojun and Zhuang, Bohan},
  booktitle={European Conference on Computer Vision},
  pages={453--470},
  year={2024},
  organization={Springer}
}

@article{shen2024longvu,
  title={Longvu: Spatiotemporal adaptive compression for long video-language understanding},
  author={Shen, Xiaoqian and Xiong, Yunyang and Zhao, Changsheng and Wu, Lemeng and Chen, Jun and Zhu, Chenchen and Liu, Zechun and Xiao, Fanyi and Varadarajan, Balakrishnan and Bordes, Florian and others},
  journal={arXiv preprint arXiv:2410.17434},
  year={2024}
}

@inproceedings{wu2019adaframe,
  title={Adaframe: Adaptive frame selection for fast video recognition},
  author={Wu, Zuxuan and Xiong, Caiming and Ma, Chih-Yao and Socher, Richard and Davis, Larry S},
  booktitle={Proceedings of the IEEE/CVF Conference on Computer Vision and Pattern Recognition},
  pages={1278--1287},
  year={2019}
}

@inproceedings{wu2019multi,
  title={Multi-agent reinforcement learning based frame sampling for effective untrimmed video recognition},
  author={Wu, Wenhao and He, Dongliang and Tan, Xiao and Chen, Shifeng and Wen, Shilei},
  booktitle={Proceedings of the IEEE/CVF International Conference on Computer Vision},
  pages={6222--6231},
  year={2019}
}

@inproceedings{korbar2019scsampler,
  title={Scsampler: Sampling salient clips from video for efficient action recognition},
  author={Korbar, Bruno and Tran, Du and Torresani, Lorenzo},
  booktitle={Proceedings of the IEEE/CVF International Conference on Computer Vision},
  pages={6232--6242},
  year={2019}
}

@inproceedings{gowda2021smart,
  title={Smart frame selection for action recognition},
  author={Gowda, Shreyank N and Rohrbach, Marcus and Sevilla-Lara, Laura},
  booktitle={Proceedings of the AAAI conference on artificial intelligence},
  volume={35},
  number={2},
  pages={1451--1459},
  year={2021}
}

@inproceedings{zhao2023search,
  title={Search-map-search: a frame selection paradigm for action recognition},
  author={Zhao, Mingjun and Yu, Yakun and Wang, Xiaoli and Yang, Lei and Niu, Di},
  booktitle={Proceedings of the IEEE/CVF Conference on Computer Vision and Pattern Recognition},
  pages={10627--10636},
  year={2023}
}

@article{yu2024frame,
  title={Frame-voyager: Learning to query frames for video large language models},
  author={Yu, Sicheng and Jin, Chengkai and Wang, Huanyu and Chen, Zhenghao and Jin, Sheng and Zuo, Zhongrong and Xu, Xiaolei and Sun, Zhenbang and Zhang, Bingni and Wu, Jiawei and others},
  journal={arXiv preprint arXiv:2410.03226},
  year={2024}
}

@inproceedings{tang2025adaptive,
  title={Adaptive keyframe sampling for long video understanding},
  author={Tang, Xi and Qiu, Jihao and Xie, Lingxi and Tian, Yunjie and Jiao, Jianbin and Ye, Qixiang},
  booktitle={Proceedings of the Computer Vision and Pattern Recognition Conference},
  pages={29118--29128},
  year={2025}
}

@article{zhu2025focus,
  title={Focus: Efficient keyframe selection for long video understanding},
  author={Zhu, Zirui and Xu, Hailun and Luo, Yang and Liu, Yong and Sarkar, Kanchan and Yang, Zhenheng and You, Yang},
  journal={arXiv preprint arXiv:2510.27280},
  year={2025}
}

@inproceedings{liu2025bolt,
  title={Bolt: Boost large vision-language model without training for long-form video understanding},
  author={Liu, Shuming and Zhao, Chen and Xu, Tianqi and Ghanem, Bernard},
  booktitle={Proceedings of the Computer Vision and Pattern Recognition Conference},
  pages={3318--3327},
  year={2025}
}

@inproceedings{fu2025video,
  title={Video-mme: The first-ever comprehensive evaluation benchmark of multi-modal llms in video analysis},
  author={Fu, Chaoyou and Dai, Yuhan and Luo, Yongdong and Li, Lei and Ren, Shuhuai and Zhang, Renrui and Wang, Zihan and Zhou, Chenyu and Shen, Yunhang and Zhang, Mengdan and others},
  booktitle={Proceedings of the IEEE/CVF conference on computer vision and pattern recognition},
  pages={24108--24118},
  year={2025}
}

@article{wu2024longvideobench,
  title={Longvideobench: A benchmark for long-context interleaved video-language understanding},
  author={Wu, Haoning and Li, Dongxu and Chen, Bei and Li, Junnan},
  journal={Advances in Neural Information Processing Systems},
  volume={37},
  pages={28828--28857},
  year={2024}
}

@article{lee2025refocus,
  title={Refocus: Reinforcement-guided frame optimization for contextual understanding},
  author={Lee, Hosu and Kim, Junho and Kim, Hyunjun and Ro, Yong Man},
  journal={arXiv preprint arXiv:2506.01274},
  year={2025}
}

@article{alayrac2022flamingo,
  title={Flamingo: a visual language model for few-shot learning},
  author={Alayrac, Jean-Baptiste and Donahue, Jeff and Luc, Pauline and Miech, Antoine and Barr, Iain and Hasson, Yana and Lenc, Karel and Mensch, Arthur and Millican, Katherine and Reynolds, Malcolm and others},
  journal={Advances in neural information processing systems},
  volume={35},
  pages={23716--23736},
  year={2022}
}

@inproceedings{li2023blip,
  title={Blip-2: Bootstrapping language-image pre-training with frozen image encoders and large language models},
  author={Li, Junnan and Li, Dongxu and Savarese, Silvio and Hoi, Steven},
  booktitle={International conference on machine learning},
  pages={19730--19742},
  year={2023},
  organization={PMLR}
}

@article{ye2024mplug,
  title={mplug-owl3: Towards long image-sequence understanding in multi-modal large language models},
  author={Ye, Jiabo and Xu, Haiyang and Liu, Haowei and Hu, Anwen and Yan, Ming and Qian, Qi and Zhang, Ji and Huang, Fei and Zhou, Jingren},
  journal={arXiv preprint arXiv:2408.04840},
  year={2024}
}

@inproceedings{liu2024tempcompass,
  title={Tempcompass: Do video llms really understand videos?},
  author={Liu, Yuanxin and Li, Shicheng and Liu, Yi and Wang, Yuxiang and Ren, Shuhuai and Li, Lei and Chen, Sishuo and Sun, Xu and Hou, Lu},
  booktitle={Findings of the Association for Computational Linguistics: ACL 2024},
  pages={8731--8772},
  year={2024}
}

@article{cai2024temporalbench,
  title={Temporalbench: Benchmarking fine-grained temporal understanding for multimodal video models},
  author={Cai, Mu and Tan, Reuben and Zhang, Jianrui and Zou, Bocheng and Zhang, Kai and Yao, Feng and Zhu, Fangrui and Gu, Jing and Zhong, Yiwu and Shang, Yuzhang and others},
  journal={arXiv preprint arXiv:2410.10818},
  year={2024}
}

@article{mangalam2023egoschema,
  title={Egoschema: A diagnostic benchmark for very long-form video language understanding},
  author={Mangalam, Karttikeya and Akshulakov, Raiymbek and Malik, Jitendra},
  journal={Advances in Neural Information Processing Systems},
  volume={36},
  pages={46212--46244},
  year={2023}
}

@inproceedings{wang2025lvbench,
  title={Lvbench: An extreme long video understanding benchmark},
  author={Wang, Weihan and He, Zehai and Hong, Wenyi and Cheng, Yean and Zhang, Xiaohan and Qi, Ji and Ding, Ming and Gu, Xiaotao and Huang, Shiyu and Xu, Bin and others},
  booktitle={Proceedings of the IEEE/CVF International Conference on Computer Vision},
  pages={22958--22967},
  year={2025}
}

@inproceedings{lei2018tvqa,
  title={Tvqa: Localized, compositional video question answering},
  author={Lei, Jie and Yu, Licheng and Bansal, Mohit and Berg, Tamara},
  booktitle={Proceedings of the 2018 conference on empirical methods in natural language processing},
  pages={1369--1379},
  year={2018}
}

@inproceedings{xiao2021next,
  title={Next-qa: Next phase of question-answering to explaining temporal actions},
  author={Xiao, Junbin and Shang, Xindi and Yao, Angela and Chua, Tat-Seng},
  booktitle={Proceedings of the IEEE/CVF conference on computer vision and pattern recognition},
  pages={9777--9786},
  year={2021}
}

@inproceedings{hu2025m,
  title={M-llm based video frame selection for efficient video understanding},
  author={Hu, Kai and Gao, Feng and Nie, Xiaohan and Zhou, Peng and Tran, Son and Neiman, Tal and Wang, Lingyun and Shah, Mubarak and Hamid, Raffay and Yin, Bing and others},
  booktitle={Proceedings of the Computer Vision and Pattern Recognition Conference},
  pages={13702--13712},
  year={2025}
}

@article{yu2023self,
  title={Self-chained image-language model for video localization and question answering},
  author={Yu, Shoubin and Cho, Jaemin and Yadav, Prateek and Bansal, Mohit},
  journal={Advances in Neural Information Processing Systems},
  volume={36},
  pages={76749--76771},
  year={2023}
}

@article{zhang2023llavar,
  title={Llavar: Enhanced visual instruction tuning for text-rich image understanding},
  author={Zhang, Yanzhe and Zhang, Ruiyi and Gu, Jiuxiang and Zhou, Yufan and Lipka, Nedim and Yang, Diyi and Sun, Tong},
  journal={arXiv preprint arXiv:2306.17107},
  year={2023}
}

@article{zhao2023svit,
  title={Svit: Scaling up visual instruction tuning},
  author={Zhao, Bo and Wu, Boya and He, Muyang and Huang, Tiejun},
  journal={arXiv preprint arXiv:2307.04087},
  year={2023}
}

@inproceedings{radford2021learning,
  title={Learning transferable visual models from natural language supervision},
  author={Radford, Alec and Kim, Jong Wook and Hallacy, Chris and Ramesh, Aditya and Goh, Gabriel and Agarwal, Sandhini and Sastry, Girish and Askell, Amanda and Mishkin, Pamela and Clark, Jack and others},
  booktitle={International conference on machine learning},
  pages={8748--8763},
  year={2021},
  organization={PmLR}
}

@inproceedings{sun2025mdp3,
  title={Mdp3: A training-free approach for list-wise frame selection in video-llms},
  author={Sun, Hui and Lu, Shiyin and Wang, Huanyu and Chen, Qing-Guo and Xu, Zhao and Luo, Weihua and Zhang, Kaifu and Li, Ming},
  booktitle={Proceedings of the IEEE/CVF International Conference on Computer Vision},
  pages={24090--24101},
  year={2025}
}

@inproceedings{zhang2025lmms,
  title={Lmms-eval: Reality check on the evaluation of large multimodal models},
  author={Zhang, Kaichen and Li, Bo and Zhang, Peiyuan and Pu, Fanyi and Cahyono, Joshua Adrian and Hu, Kairui and Liu, Shuai and Zhang, Yuanhan and Yang, Jingkang and Li, Chunyuan and others},
  booktitle={Findings of the Association for Computational Linguistics: NAACL 2025},
  pages={881--916},
  year={2025}
}

@article{bai2025qwen3,
  title={Qwen3-vl technical report},
  author={Bai, Shuai and Cai, Yuxuan and Chen, Ruizhe and Chen, Keqin and Chen, Xionghui and Cheng, Zesen and Deng, Lianghao and Ding, Wei and Gao, Chang and Ge, Chunjiang and others},
  journal={arXiv preprint arXiv:2511.21631},
  year={2025}
}

@article{bai2025qwen2,
  title={Qwen2.5-vl technical report},
  author={Bai, Shuai and Chen, Keqin and Liu, Xuejing and Wang, Jialin and Ge, Wenbin and Song, Sibo and Dang, Kai and Wang, Peng and Wang, Shijie and Tang, Jun and others},
  journal={arXiv preprint arXiv:2502.13923},
  year={2025}
}

@misc{openai2024gpt4o,
  author    = {\vspace{0mm}OpenAI},
  title     = {Hello GPT-4o},
  howpublished = {\url{https://openai.com/index/hello-gpt-4o/}},
  year      = {2024}
}

@article{team2023gemini,
  title={Gemini: a family of highly capable multimodal models},
  author={Team, Gemini and Anil, Rohan and Borgeaud, Sebastian and Alayrac, Jean-Baptiste and Yu, Jiahui and Soricut, Radu and Schalkwyk, Johan and Dai, Andrew M and Hauth, Anja and Millican, Katie and others},
  journal={arXiv preprint arXiv:2312.11805},
  year={2023}
}

@article{zhang2024llava,
  title={Llava-video: Video instruction tuning with synthetic data},
  author={Zhang, Yuanhan and Wu, Jinming and Li, Wei and Li, Bo and Ma, Zejun and Liu, Ziwei and Li, Chunyuan},
  journal={arXiv preprint arXiv:2410.02713},
  year={2024}
}

@article{wang2025internvl3,
  title={Internvl3.5: Advancing open-source multimodal models in versatility, reasoning, and efficiency},
  author={Wang, Weiyun and Gao, Zhangwei and Gu, Lixin and Pu, Hengjun and Cui, Long and Wei, Xingguang and Liu, Zhaoyang and Jing, Linglin and Ye, Shenglong and Shao, Jie and others},
  journal={arXiv preprint arXiv:2508.18265},
  year={2025}
}

@article{yu2025minicpm,
  title={Minicpm-v 4.5: Cooking efficient mllms via architecture, data, and training recipe},
  author={Yu, Tianyu and Wang, Zefan and Wang, Chongyi and Huang, Fuwei and Ma, Wenshuo and He, Zhihui and Cai, Tianchi and Chen, Weize and Huang, Yuxiang and Zhao, Yuanqian and others},
  journal={arXiv preprint arXiv:2509.18154},
  year={2025}
}

@inproceedings{li2024llama,
  title={Llama-vid: An image is worth 2 tokens in large language models},
  author={Li, Yanwei and Wang, Chengyao and Jia, Jiaya},
  booktitle={European Conference on Computer Vision},
  pages={323--340},
  year={2024},
  organization={Springer}
}

@inproceedings{shu2025video,
  title={Video-xl: Extra-long vision language model for hour-scale video understanding},
  author={Shu, Yan and Liu, Zheng and Zhang, Peitian and Qin, Minghao and Zhou, Junjie and Liang, Zhengyang and Huang, Tiejun and Zhao, Bo},
  booktitle={Proceedings of the Computer Vision and Pattern Recognition Conference},
  pages={26160--26169},
  year={2025}
}

@article{lan2024vidcompress,
  title={Vidcompress: Memory-enhanced temporal compression for video understanding in large language models},
  author={Lan, Xiaohan and Yuan, Yitian and Jie, Zequn and Ma, Lin},
  journal={arXiv preprint arXiv:2410.11417},
  year={2024}
}

@inproceedings{huang2025prunevid,
  title={Prunevid: Visual token pruning for efficient video large language models},
  author={Huang, Xiaohu and Zhou, Hao and Han, Kai},
  booktitle={Findings of the Association for Computational Linguistics: ACL 2025},
  pages={19959--19973},
  year={2025}
}

@inproceedings{li2024vidtome,
  title={Vidtome: Video token merging for zero-shot video editing},
  author={Li, Xirui and Ma, Chao and Yang, Xiaokang and Yang, Ming-Hsuan},
  booktitle={Proceedings of the IEEE/CVF Conference on Computer Vision and Pattern Recognition},
  pages={7486--7495},
  year={2024}
}

@article{luo2026video,
  title={Video-rag: Visually-aligned retrieval-augmented long video comprehension},
  author={Luo, Yongdong and Zheng, Xiawu and Li, Guilin and Yin, Shukang and Lin, Haojia and Fu, Chaoyou and Huang, Jinfa and Ji, Jiayi and Chao, Fei and Luo, Jiebo and others},
  journal={Advances in Neural Information Processing Systems},
  volume={38},
  pages={168008--168033},
  year={2026}
}

@article{yuan2025memory,
  title={Memory-enhanced retrieval augmentation for long video understanding},
  author={Yuan, Huaying and Liu, Zheng and Qin, Minghao and Qian, Hongjin and Shu, Yan and Dou, Zhicheng and Wen, Ji-Rong and Sebe, Nicu},
  journal={arXiv preprint arXiv:2503.09149},
  year={2025}
}

@inproceedings{wang2025videotree,
  title={Videotree: Adaptive tree-based video representation for llm reasoning on long videos},
  author={Wang, Ziyang and Yu, Shoubin and Stengel-Eskin, Elias and Yoon, Jaehong and Cheng, Feng and Bertasius, Gedas and Bansal, Mohit},
  booktitle={Proceedings of the Computer Vision and Pattern Recognition Conference},
  pages={3272--3283},
  year={2025}
}

@inproceedings{ma2025drvideo,
  title={Drvideo: Document retrieval based long video understanding},
  author={Ma, Ziyu and Gou, Chenhui and Shi, Hengcan and Sun, Bin and Li, Shutao and Rezatofighi, Hamid and Cai, Jianfei},
  booktitle={Proceedings of the Computer Vision and Pattern Recognition Conference},
  pages={18936--18946},
  year={2025}
}

@inproceedings{li2023tcovis,
  title={Tcovis: Temporally consistent online video instance segmentation},
  author={Li, Junlong and Yu, Bingyao and Rao, Yongming and Zhou, Jie and Lu, Jiwen},
  booktitle={Proceedings of the IEEE/CVF International Conference on Computer Vision},
  pages={1097--1107},
  year={2023}
}

@inproceedings{yang2025egolife,
  title={Egolife: Towards egocentric life assistant},
  author={Yang, Jingkang and Liu, Shuai and Guo, Hongming and Dong, Yuhao and Zhang, Xiamengwei and Zhang, Sicheng and Wang, Pengyun and Zhou, Zitang and Xie, Binzhu and Wang, Ziyue and others},
  booktitle={Proceedings of the Computer Vision and Pattern Recognition Conference},
  pages={28885--28900},
  year={2025}
}

@article{hu2025emobench,
  title={Emobench-m: Benchmarking emotional intelligence for multimodal large language models},
  author={Hu, He and You, Lianzhong and Xu, Hongbo and Wang, Qianning and Yu, Fei Richard and Ma, Fei and Cheng, Zebang and Lian, Zheng and Zhou, Yucheng and Cui, Laizhong},
  journal={arXiv preprint arXiv:2502.04424},
  year={2025}
}

@inproceedings{zhou2018towards,
  title={Towards automatic learning of procedures from web instructional videos},
  author={Zhou, Luowei and Xu, Chenliang and Corso, Jason},
  booktitle={Proceedings of the AAAI conference on artificial intelligence},
  volume={32},
  number={1},
  year={2018}
}

@inproceedings{xu2025multiagentesc,
  title={Multiagentesc: A llm-based multi-agent collaboration framework for emotional support conversation},
  author={Xu, Yangyang and Hu, Jinpeng and Zhao, Zhuoer and Duan, Zhangling and Sun, Xiao and Yang, Xun},
  booktitle={Proceedings of the 2025 Conference on Empirical Methods in Natural Language Processing},
  pages={4665--4681},
  year={2025}
}

@inproceedings{zhang2025towards,
  title={Towards video thinking test: A holistic benchmark for advanced video reasoning and understanding},
  author={Zhang, Yuanhan and Chew, Yunice and Dong, Yuhao and Leo, Aria and Hu, Bo and Liu, Ziwei},
  booktitle={Proceedings of the IEEE/CVF International Conference on Computer Vision},
  pages={20626--20636},
  year={2025}
}

@article{team2026script,
  title={Script-a-video: Deep structured audio-visual captions via factorized streams and relational grounding},
  author={Team, Tencent Hunyuan},
  journal={arXiv preprint arXiv:2604.11244},
  year={2026}
}

@article{romero2023zelda,
  title={Zelda: Video analytics using vision-language models},
  author={Romero, Francisco and Winston, Caleb and Hauswald, Johann and Zaharia, Matei and Kozyrakis, Christos},
  journal={arXiv preprint arXiv:2305.03785},
  year={2023}
}

@inproceedings{li2026divide,
  title={Divide, then ground: Adapting frame selection to query types for long-form video understanding},
  author={Li, Jialuo and Li, Bin and Li, Jiahao and Lu, Yan},
  booktitle={Proceedings of the IEEE/CVF Conference on Computer Vision and Pattern Recognition},
  pages={11369--11380},
  year={2026}
}

@inproceedings{zou2026air,
  title={Air: Enabling adaptive, iterative, and reasoning-based frame selection for video question answering},
  author={Zou, Yuanhao and Jin, Shengji and Deng, Andong and Zhao, Youpeng and Wang, Jun and Chen, Chen},
  booktitle={International Conference on Learning Representations},
  volume={2026},
  pages={11302--11329},
  year={2026}
}

@inproceedings{hu2025beyond,
  title={Beyond emotion recognition: A multi-turn multimodal emotion understanding and reasoning benchmark},
  author={Hu, Jinpeng and Shi, Hongchang and Dai, Chongyuan and Li, Zhuo and Song, Peipei and Wang, Meng},
  booktitle={Proceedings of the 33rd ACM International Conference on Multimedia},
  pages={5814--5823},
  year={2025}
}

@article{hu2026mmhbench,
  title={MMHBench: A Multi-Perspective Benchmark for Mental Health Understanding in Long-Form Videos},
  author={Hu, Jinpeng and Wang, Erqiang and Wang, Shan and Li, Zhuo and Song, Peipei and Yang, Xun and Wang, Meng},
  journal={arXiv preprint arXiv:2607.27895},
  year={2026}
}

@inproceedings{dai2026psyche,
  title={Psyche-r1: Towards reliable psychological llms through unified empathy, expertise, and reasoning},
  author={Dai, Chongyuan and Hu, Jinpeng and Shi, Hongchang and Li, Zhuo and Guo, Dan and Yang, Xun and Wang, Meng},
  booktitle={Proceedings of the 64th Annual Meeting of the Association for Computational Linguistics (Volume 1: Long Papers)},
  pages={24889--24906},
  year={2026}
}

@inproceedings{hu2026agentmental,
  title={Agentmental: An interactive multi-agent framework for explainable and adaptive mental health assessment},
  author={Hu, Jinpeng and Wang, Ao and Xie, Qianqian and Li, Zhuo and Ma, Hui and Guo, Dan},
  booktitle={Proceedings of the AAAI Conference on Artificial Intelligence},
  volume={40},
  number={37},
  pages={31050--31058},
  year={2026}
}
}


\end{document}